\documentclass[11pt, letterpaper]{template}
\usepackage[authoryear, round]{natbib}

\usepackage{subcaption}
\usepackage{calc}
\usepackage[export]{adjustbox}
\usepackage[version=4]{mhchem}
\usepackage{multirow}
\usepackage[T1]{fontenc}
\usepackage{lmodern}

\usepackage{longtable}
\usepackage{framed}
\definecolor{shadecolor}{RGB}{248,248,248}

\usepackage[capitalize,noabbrev]{cleveref}

\usepackage[dvipsnames]{xcolor}
\hypersetup{
    colorlinks=true,
    linkcolor=Cyan,
    citecolor=Green,
    filecolor=magenta,
    urlcolor=magenta,
}

\theoremstyle{plain}

\theoremstyle{definition}

\theoremstyle{remark}

\usepackage[textsize=tiny]{todonotes}

\usepackage{listings}
\usepackage[most]{tcolorbox}
\definecolor{NavyBlue}{rgb}{0.0, 0.0, 0.5}
\definecolor{DarkerNavy}{rgb}{0.0, 0.05, 0.3}
\definecolor{DeepNavy}{rgb}{0.0, 0.0, 0.4}
\definecolor{DustyBlue}{rgb}{0.2, 0.3, 0.4}
\tcbset{
  promptbox/.style={
    top=5pt,
    bottom=5pt,
    colback=lightgray!20,
    colframe=black,
    colbacktitle=DustyBlue,
    enhanced,
    center,
    attach boxed title to top center={yshift=-0.1in,xshift=0.0in},
    boxed title style={boxrule=0pt,colframe=white,},
    before upper={\parindent0pt\setlength{\parskip}{2pt}},
  }
}
\newtcolorbox{promptbox}[2][]{promptbox, title=#2,#1}

\newtcbtheorem[number within=section]{question}{Question}{
    top=10pt,
    colback=lightgray!20,
    colframe=Black,
    colbacktitle=DustyBlue,
    enhanced,
    center,
    attach boxed title to top center={yshift=-0.1in,xshift=0.0in},
    boxed title style={boxrule=0pt,colframe=white,},
    breakable
}{qn}

\newtcbtheorem[number within=section]{prompt}{Prompt}{
    top=10pt,
    colback=lightgray!20,
    colframe=Black,
    colbacktitle=Cerulean,
    enhanced,
    center,
    attach boxed title to top center={yshift=-0.1in,xshift=0.0in},
    boxed title style={boxrule=0pt,colframe=white,},
    breakable
}{pp}

\newtcbtheorem[number within=section]{case}{Case}{
    top=10pt,
    colback=lightgray!20,
    colframe=Black,
    colbacktitle=Tan,
    enhanced,
    center,
    attach boxed title to top center={yshift=-0.1in,xshift=0.0in},
    boxed title style={boxrule=0pt,colframe=white,},
    breakable
}{cs}

\usepackage{xspace}
\newcommand{\name}{\texttt{InternVideo3}\xspace}

\newcommand{\blfootnote}[1]{%
  \begingroup
  \renewcommand\thefootnote{}\footnote{#1}%
  \addtocounter{footnote}{-1}%
  \endgroup
}

\usepackage{pifont}

\title{\name: Agentify Foundation Models with Multimodal Contextual Reasoning}

\author[1*]{Ziang Yan} \author[2,3*]{Sheng Xia} \author[1*]{Jiashuo Yu} \author[1*]{Yue Wu} \author[1]{Tianxiang Jiang} \author[1]{Songze Li} \author[1]{Kanghui Tian} \author[1]{Yicheng Xu} \author[1]{Yinan He} \author[1]{Kai Chen} \author[3,1]{Limin Wang} \author[1]{Yu Qiao} \author[1$\dagger$]{Yi Wang}
\affil[1]{Shanghai AI Laboratory} \affil[2]{Shanghai Innovation Institute}
\affil[3]{Nanjing University}
\newcommand{\projectpage}{\textbf{Project Page:} \url{https://github.com/OpenGVLab/InternVideo/tree/main/InternVideo3}}

\begin{document}

\begin{abstract}
Recent progress in foundation models has increasingly shifted from one-shot prediction toward \emph{agentic} behavior, where models solve tasks through multi-step reasoning, tool use, memory, and self-correction. However, much of the open-source momentum has centered on text-dominant settings such as coding, search, and long-context tool use, while \emph{long-horizon multimodal} tasks remain comparatively underexplored. This gap is especially visible in video, where real-world tasks often require sustained temporal understanding, visually grounded evidence gathering, and iterative interaction with external tools or memory rather than a single-pass question-answering step.
We present \name, a framework for improving such capabilities through \emph{Multimodal Contextual Reasoning} (MCR), a formulation that treats multimodal understanding as a closed-loop process over a shared evolving context. In MCR, multimodal observations, instructions, intermediate reasoning, tool actions, feedback, and memory are all represented within a unified context that is updated over time. This makes long-video understanding a process of evidence accumulation, belief revision, and verification, and provides a practical abstraction for multimodal agentic behavior.
To make such long-horizon rollouts efficient, we introduce \emph{Multimodal Multi-head Latent Attention} (M$^2$LA), a token-preserving attention reparameterization that compresses KV-cache states while retaining the full multimodal token stream. We further develop a staged training recipe consisting of continued pretraining after M$^2$LA conversion, short-to-long supervised fine-tuning for video, rule-based reinforcement learning on verifiable tasks, and on-policy distillation from stronger teacher models.
Experiments on short-video and long-video benchmarks show that \name\ achieves strong performance among open video models, with especially notable gains on long-horizon tasks such as Video-MME, MLVU, and EgoSchema. We also instantiate the model as a video agent with retrieval and verification tools, illustrating how recursive multimodal reasoning can support more robust evidence-grounded behavior. Overall, our results suggest that efficient context handling and closed-loop multimodal reasoning are important ingredients for adapting open multimodal models toward long-horizon visually grounded agency.

\end{abstract}

\blfootnote{*Equal contribution. $\dagger$Corresponding author.}

\maketitle

\section{Introduction}

Foundation models are increasingly evaluated not only by how well they answer questions, but by how well they \emph{act} over extended horizons. Across recent research, two themes have become especially prominent: \emph{agents}, which emphasize multi-step reasoning~\citep{react}, tool use~\citep{schick2023toolformer}, memory~\citep{park2023generative}, and self-correction~\citep{shinn2023reflexion}; and \emph{world modeling}, which emphasizes persistent internal state, temporal abstraction, and reasoning over evolving environments \citep{ha2018world}. Although these directions are often studied separately, they share a common shift in perspective: moving from one-shot prediction toward systems that maintain, update, and act on contextual knowledge over time.

So far, however, much of the open-source progress in agentic AI has concentrated on \emph{text-dominant} settings, including coding agents~\citep{yang2024swe, wang2025openhands}, browser~\citep{yao2022webshop,zhou2024webarena} and search agents~\citep{react,jin2025search,zheng2025deepresearcher}, function calling~\citep{schick2023toolformer, qin2024toolllm}, and ultra-long-context language tasks~\citep{zhang2024bench, bai2025longbench}. By comparison, \emph{visually grounded long-horizon reasoning} remains significantly less developed. Existing multimodal large language models (MLLMs)~\citep{qwen25vl,qwen3vl,internvl35,li2024llava} have made impressive progress on image understanding and short-video QA, but they still struggle on tasks that require sustained temporal understanding, evidence localization, spatial reasoning, iterative evidence gathering, and interaction with external tools or memory. Yet these are precisely the settings in which multimodal agents must operate: the model must determine what it has seen, what remains uncertain, what additional evidence is needed, and whether its current conclusion is sufficiently grounded to answer or act.

This gap matters for both practical and scientific reasons. Practically, many real workloads, including long-form video analysis~\citep{longvideobench,wang2024lvbench}, instructional understanding~\citep{howto100m}, surveillance review~\citep{sultani2018real}, egocentric perception~\citep{ego4d,egoschema}, and evidence-grounded temporal reasoning~\citep{lei2018tvqa,nextgqa,liu2024tempcompass}, cannot be reliably solved by single-pass multimodal QA. They require revisiting observations, preserving relevant memory, localizing supporting evidence, invoking specialized tools, and revising conclusions when new information arrives. Scientifically, these tasks offer a concrete testbed for capabilities often associated with multimodal agency and world-model-like reasoning: latent state tracking~\citep{hansen2022temporal,hansen2024td}, temporal abstraction~\citep{machado2023temporal, kong2024latent}, uncertainty-aware evidence selection~\citep{asai2024self}, and closed-loop interaction between perception and decision making~\citep{ahn2022can,brohan2022rt}. In this sense, long-horizon video reasoning is not merely another benchmark category, but an important substrate for studying visually grounded agency in foundation models.

In this work, we study how to move open multimodal models \emph{toward} this regime by \emph{agentifying their reasoning process} rather than assuming a one-shot mapping from video to answer. We propose \emph{Multimodal Contextual Reasoning} (MCR), a unified formulation in which multimodal observations, task instructions, intermediate reasoning traces, tool actions, feedback, and memory are all represented within a shared evolving context. Under MCR, long-video understanding becomes a recursive process of \emph{observe, reason, act, receive feedback, and update context}. This formulation does not attempt to build a full action-conditioned world simulator. Instead, it provides a practical way for a multimodal foundation model to maintain and refine a task-relevant contextual belief state while reasoning over long visual streams and interacting with tools.

A major obstacle to this paradigm is efficiency. In long multimodal rollouts, the bottleneck is not only the number of visual tokens, but the growing KV-cache required to preserve observations, reasoning traces, tool outputs, and memory across many decoding steps. Existing approaches often reduce cost by aggressive frame subsampling, retrieval, or summarization. While useful, such methods may discard information that later becomes important for reasoning or verification. We therefore pursue a complementary direction: preserving the full multimodal token stream while reducing memory usage \emph{inside} attention. To this end, we introduce \emph{Multimodal Multi-head Latent Attention} (M$^2$LA), a long-context-efficient attention design that compresses cached KV states while reconstructing head-specific representations on the fly. This makes longer multimodal rollouts feasible under practical hardware constraints.

Building on this efficient attention foundation, we develop a staged training recipe tailored to long-horizon multimodal reasoning. Starting from an open multimodal backbone, we first perform continued pretraining after M$^2$LA conversion to recover language capability and multimodal alignment under the new attention parameterization. We then carry out large-scale long-video supervised fine-tuning with a short-to-long curriculum, followed by rule-based reinforcement learning on verifiable tasks and on-policy distillation~\citep{agarwal2024policy,lu2025onpolicydistillation,xu2025speculative} from stronger teacher models. This pipeline strengthens the model's ability to reason over dense visual evidence and extended temporal dependencies without requiring frontier-scale base models.

Empirically, \name\ achieves strong results across short-video, long-video, and spatiotemporal reasoning benchmarks, with especially notable gains on long-horizon tasks such as Video-MME~\citep{videomme}, MLVU~\citep{mlvu}, and EgoSchema~\citep{egoschema}. We further instantiate the model as a video agent with retrieval and verification tools, illustrating how recursive multimodal reasoning can support more robust evidence-grounded behavior. Taken together, our results suggest that \emph{context efficiency} and \emph{closed-loop multimodal reasoning} are useful ingredients for adapting open multimodal models toward long-horizon visually grounded agency.

Our contributions are summarized as follows:
\begin{itemize}[leftmargin=*]
    \item We propose \textbf{Multimodal Contextual Reasoning (MCR)}, a unified formulation for long-horizon multimodal reasoning in which observations, intermediate reasoning, tool use, feedback, and memory are represented within a shared evolving context.
    \item We introduce \textbf{M$^2$LA}, a long-context-efficient attention reparameterization that compresses KV-cache states while preserving the full multimodal token stream, enabling longer multimodal rollouts under constrained hardware budgets.
    \item We develop a practical training recipe for long-video reasoning, combining continued pretraining after attention conversion, short-to-long supervised fine-tuning, rule-based reinforcement learning on verifiable tasks, and on-policy distillation from stronger teacher models.
    \item We demonstrate strong results on short-video, long-video, and spatiotemporal reasoning benchmarks, and further present a video-agent instantiation that illustrates the practical value of recursive multimodal reasoning for evidence-grounded behavior.
\end{itemize}

\section{Related Work}

\paragraph{Multimodal Large Language Models.}
The rapid development of multimodal large language models has expanded foundation-model capabilities from image-text alignment and image understanding \citep{clip,flamingo,llava,pmlr-v202-li23q,internvl,internvl2} to more general video understanding and multimodal reasoning \citep{internvideo,internvideo2,wang2025internvideo2,videollama2,Llava-onevision,qwen3vl}. Recent open-source video MLLMs achieve strong performance on standard short-to-medium video benchmarks such as VideoMME \citep{videomme}, MVBench \citep{mvbench}, LVBench \citep{wang2024lvbench}, and related instruction-following settings. However, performance still degrades on genuinely long videos or tasks requiring sustained temporal reasoning, evidence localization, and multi-step context accumulation. A common limitation is that multimodal inputs are often treated as passive context to be summarized once, rather than as evolving evidence to be revisited and refined during reasoning. Our work focuses on this longer-horizon regime, where the challenge is not only to perceive visual content, but to maintain and update a coherent multimodal context over time.

\paragraph{Agents.}
Turning foundation models into agents has become a major research direction since ReAct \citep{react}, which showed the effectiveness of interleaving reasoning and action. Subsequent work extend this paradigm to program synthesis, tool use, GUI interaction, mobile interfaces, and multimodal environments, including ViperGPT \citep{vipergpt}, VisProg \citep{visprog}, CogAgent \citep{cogagent}, AppAgent \citep{appagent}, Mobile-Agent \citep{mobileagent}, and a growing family of coding and search agents~\citep{yang2024swe, wang2025openhands, jin2025search, zheng2025deepresearcher}. In multimodal settings, the core challenge is often not only selecting the next tool, but deciding what perceptual evidence is missing, where to look next, and whether current evidence is sufficient to answer or act. Prior video-agent systems \citep{videoagent1,videoagent2} demonstrate the value of tool-augmented video reasoning, such as frame sampling, temporal grounding, or subtitle retrieval, but they often rely on planner-controller decompositions, short interaction horizons, or task-specific pipelines. Our work instead studies how to support longer-horizon visually grounded reasoning within a unified MLLM context, where observations, intermediate reasoning, tool outputs, and memory are all represented within a common rollout.

\paragraph{World Models and Predictive Video Learning.}
World models aim to learn compact internal state representations that support prediction, planning, and decision making in partially observed environments. This line of work spans latent dynamics models for reinforcement learning \citep{ha2018world,hafner2019dream,vjepa2}, action-conditioned predictive models, generative simulators, and recent self-supervised predictive representation learning approaches. An influential direction is the \emph{Joint-Embedding Predictive Architecture} (JEPA), learning by predicting missing or future content in a learned embedding space rather than reconstructing pixels \citep{ijepa,vjepa,vjepa2,vjepa21}. JEPA-style methods emphasize learning abstract, predictable, and task-relevant structure while discarding low-level detail unnecessary for downstream reasoning or control. This paradigm has evolved from image representation learning in I-JEPA \citep{ijepa} to video predictive learning in V-JEPA \citep{vjepa}, and more recently to V-JEPA2 \citep{vjepa2}, connecting large-scale video pretraining with understanding, prediction, and robot planning. These works suggest predictive latent representations are a promising substrate for world understanding and, with additional action supervision, for planning. Our work is complementary to this literature. Rather than training a full action-conditioned latent world model from interaction data, we study how a multimodal foundation model can maintain and update a task-relevant \emph{contextual belief state} over long visual streams during reasoning and tool interaction.

\paragraph{Context Modeling and Engineering.}
Handling long multimodal sequences remains a central challenge in both LLMs and MLLMs. Existing strategies include sparse frame sampling, hierarchical temporal modeling, retrieval-augmented frame selection, summarization, memory tokens, and context compression. For long videos and long documents, systems such as MovieChat \citep{moviechat}, LongVU \citep{longvu}, and Gemini 1.5 \citep{gemini} employ retrieval, summarization, or compression to fit large evidence sets into finite contexts. Meanwhile, recent LLMs have increasingly focused on ultra-long contexts for agentic workloads \citep{deepseekai2026deepseekv4}. However, multimodal agency requires more than simply fitting long context into memory: it requires dynamic, reasoning-guided evidence selection and the preservation of causal dependencies across observations, actions, and feedback. Our M$^2$LA architecture addresses this challenge from an attention-efficiency perspective by compressing KV-cache states without dropping tokens, while MCR provides a formulation for recursively updating multimodal context during long-horizon rollouts.

\paragraph{Reasoning, Reflection, and Test-Time Interaction.}
Enhancing reasoning in foundation models has progressed through prompting techniques such as Chain-of-Thought \citep{cot}, Tree-of-Thought \citep{tot}, self-reflection, and through learning-based methods such as process supervision, outcome-supervised RL, and test-time scaling. In multimodal settings, these mechanisms become tightly coupled with perception: the model must decide whether current observations suffice, whether more evidence is needed, and which perception or tool operation should be performed next. Visual search and grounding methods such as V* \citep{vstar} suggest that reasoning can guide what evidence to inspect within an image or video. Our work follows this direction, but places reasoning, perception updates, tool feedback, and memory into a single evolving multimodal context. This enables a closed-loop formulation in which belief updates and evidence gathering are part of the same sequence model, rather than auxiliary stages outside it.

\section{Method}
\label{sec:method}

\begin{figure}[t]
    \centering
    \includegraphics[width=\linewidth]{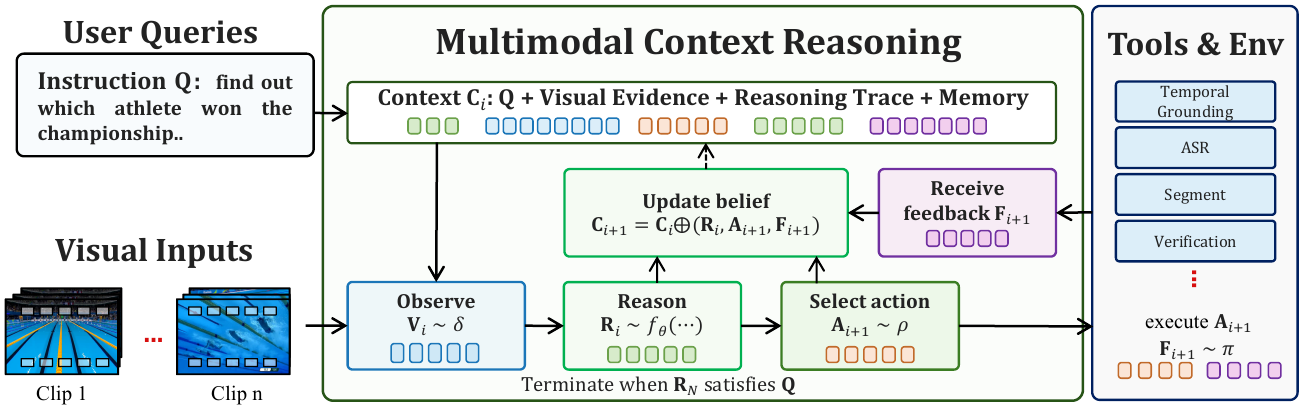}
    \caption{Framework of MCR.}
    \label{fig:mcr}
\end{figure}

We present \name, a framework for improving long-horizon multimodal reasoning through three complementary components: 
(1) \emph{Multimodal Contextual Reasoning} (MCR), a formulation that represents observations, intermediate reasoning, tool actions, feedback, and memory within a shared evolving context; 
(2) \emph{Multimodal Multi-head Latent Attention} (M$^2$LA), an efficient attention reparameterization that reduces KV-cache cost for long multimodal rollouts; and 
(3) a staged training recipe that restores pretrained capability after attention conversion and then specializes the model for long-video reasoning. 

The central idea is to treat long-horizon multimodal understanding not as a one-pass mapping from video to answer, but as a closed-loop process in which the model repeatedly observes, reasons, acts, receives feedback, and updates its contextual state. In this section, we first introduce the MCR formulation, then describe the long-context-efficient attention design that makes such rollouts practical, followed by our training recipe and video-agent instantiation.

\subsection{Multimodal Contextual Reasoning}

\paragraph{Motivation.}
Conventional multimodal QA typically assumes that a model receives a fixed image or video and produces an answer in a single forward pass. This abstraction is effective for short and self-contained examples, but becomes increasingly inadequate for long videos and visually grounded agentic tasks. In these settings, the model must often decide what evidence is already available, what information remains missing, whether to inspect another temporal segment or spatial region, whether to invoke an external tool, and whether its current conclusion should be revised in light of new evidence. These requirements are more naturally captured by an \emph{evolving contextual state} than by a single static prompt.

We therefore formulate multimodal reasoning as \textbf{Multimodal Contextual Reasoning (MCR)}, in which task-relevant information, including multimodal observations, user instructions, intermediate reasoning traces, tool actions, external feedback, and memory, is represented within a shared context that grows over time. This context acts as a task-grounded belief state: it records what the model has observed, what it has inferred, what actions it has taken, and what uncertainty remains.

\paragraph{Closed-Loop Rollout.}
Given a user query $\mathbf{Q}$ and an initial multimodal observation $\mathbf{V}_{\texttt{init}}$, MCR proceeds through a multi-step rollout. At step $i$, the model conditions on the current visual evidence $\mathbf{V}_i$, the current action trace $\mathbf{A}_i$, the tool or environment feedback $\mathbf{F}_i$, and the accumulated context $\mathbf{C}_i$, and produces an intermediate reasoning state or response $\mathbf{R}_i$:
\begin{align}
\mathbf{R}_i \sim f_{\theta}\!\left(\mathbf{Q}\ \middle|\ \mathbf{V}_i,\ \mathbf{A}_i,\ \mathbf{F}_i,\ \mathbf{C}_i\right), \qquad i = 1,2,\dots,N .
\label{eq:mcr_main_rewrite}
\end{align}
The resulting reasoning state triggers an updated action decision, tool call, or perception request:
\begin{align}
\mathbf{A}_{i+1} &\sim \rho\!\left(\mathbf{A}_i\ \middle|\ \mathbf{R}_i,\mathbf{F}_i,\mathbf{Q}\right), \\
\mathbf{F}_{i+1} &\sim \pi\!\left(\mathbf{A}_{i+1}\ \middle|\ \mathbf{R}_i,\mathbf{Q}\right), \\
\mathbf{V}_{i+1} &\sim \delta\!\left(\mathbf{V}_i\ \middle|\mathbf{C}_{i+1}\right),
\label{eq:mcr_transition_rewrite}
\end{align}
where $\rho(\cdot)$ denotes the policy that selects the next action, $\pi(\cdot)$ denotes environment or tool execution, and $\delta(\cdot)$ denotes the perception update operator that determines what visual evidence should be observed next. Note the newly produced reasoning, action, and feedback are appended to the context:
\begin{align}
\mathbf{C}_{i+1} = \mathbf{C}_i \oplus \left(\mathbf{R}_i,\mathbf{A}_{i+1},\mathbf{F}_{i+1}\right),
\qquad
\mathbf{C}_1 = \mathbf{T}_{\texttt{sys}},
\qquad
(\mathbf{V}_1,\mathbf{A}_1,\mathbf{F}_1)=(\mathbf{V}_{\texttt{init}},\varnothing,\varnothing),
\label{eq:mcr_context_rewrite}
\end{align}
where $\oplus$ denotes context aggregation, optionally combined with summarization or compression to control growth over long rollouts. The rollout terminates after $N$ steps once the model emits a final answer or a termination action:
$\mathbf{R}_N \models \mathbf{Q}$.

In practice, we implement this process as an autoregressive rollout in which the model alternates between generating intermediate reasoning tokens, tool calls, and final answers, while appending tool outputs and memory summaries back into the same multimodal context.

\paragraph{Interpretation.}
MCR can be understood as a \emph{contextual belief-update process} implemented inside an MLLM. Rather than learning a standalone action-conditioned world simulator, the model incrementally constructs and revises a task-relevant contextual state through multimodal evidence accumulation, reasoning, and tool interaction. Our goal is therefore not full predictive world modeling, but a practical mechanism for maintaining and updating grounded beliefs over long visual streams. In this sense, MCR is best viewed as a useful abstraction for long-horizon multimodal reasoning and visually grounded agentic behavior.

\paragraph{Actions and Tools in MCR.}
In our instantiation, actions may include:
\begin{itemize}[leftmargin=*]
    \item \textbf{Perceptive actions}, such as requesting a temporal zoom-in, revisiting a segment, or selecting a more informative clip.
    \item \textbf{Tool actions}, such as invoking ASR, segmentation, temporal grounding, or web search.
    \item \textbf{Memory actions}, such as reading, writing, or summarizing intermediate context.
    \item \textbf{Verification actions}, such as checking whether sufficient evidence supports the current answer.
    \item \textbf{Termination actions}, such as returning the final response.
\end{itemize}
This unified view allows diverse agentic behaviors to be represented within the same sequence modeling framework.

\subsection{Long-Context Efficient Attention with M$^2$LA}

MCR places observations, reasoning traces, tool outputs, and memory updates into a shared evolving context. As the rollout grows, the main bottleneck is not only the number of input tokens, but the \emph{KV-cache footprint} required to preserve this context during decoding. This issue is especially severe in multimodal settings, where visual evidence already occupies a large fraction of the context and additional agent-like traces accumulate over multiple reasoning steps.

\begin{figure}[t]
    \centering
    \includegraphics[width=\linewidth]{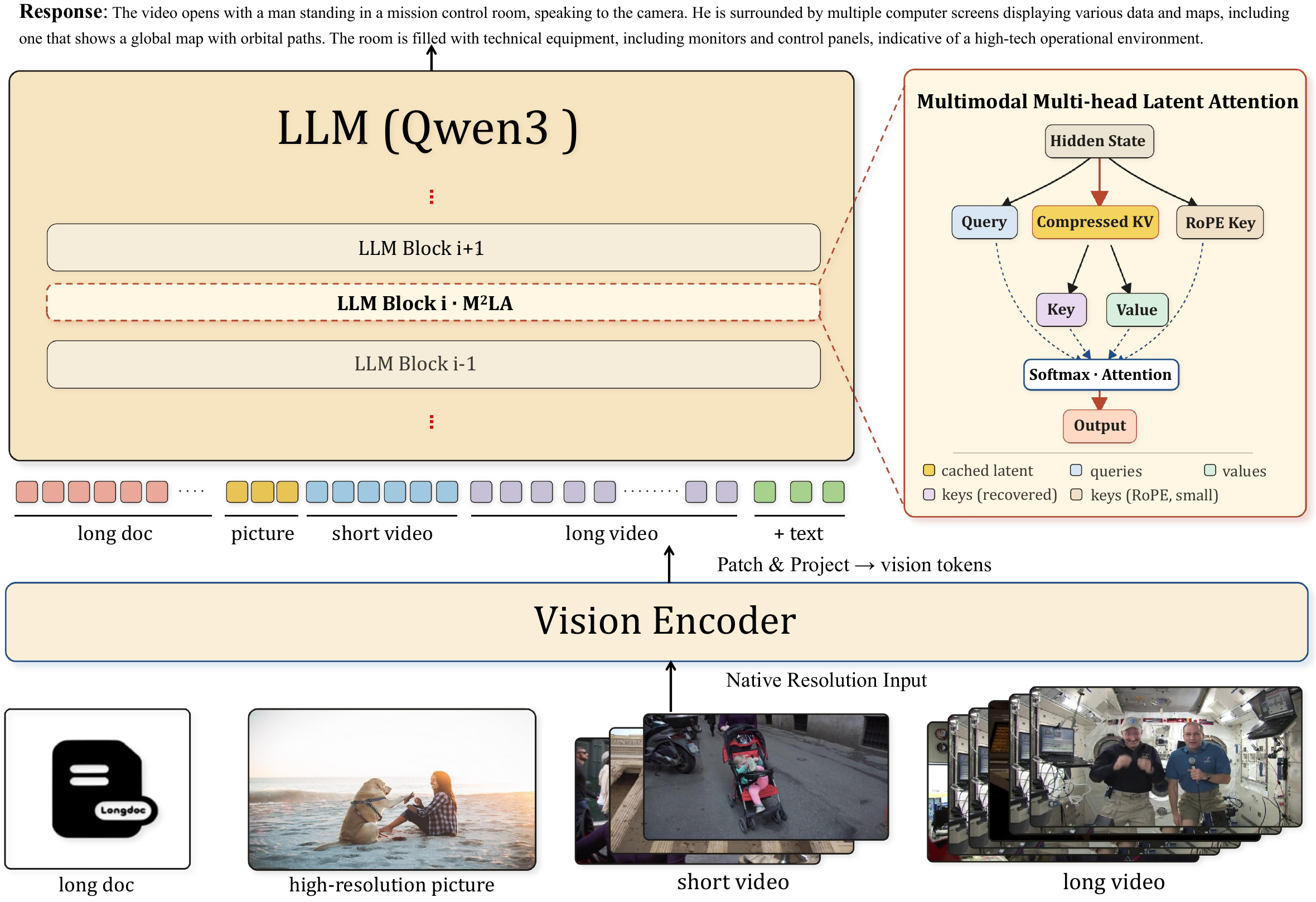}
    \caption{Architecture of \name.}
    \label{fig:method_overview}
\end{figure}

A straightforward way to reduce this cost is to drop tokens through subsampling, retrieval, or summarization. While useful, such methods may remove information that later becomes important for verification or evidence integration. Since our goal is to preserve as much multimodal evidence as possible while still enabling long-horizon rollouts, we pursue a complementary approach: \emph{compressing cached attention states rather than dropping tokens}. To this end, we introduce \textbf{Multimodal Multi-head Latent Attention} (M$^2$LA), shown in Figure~\ref{fig:method_overview}.

M$^2$LA replaces standard GQA-style~\citep{ainslie2023gqa} attention blocks with a latent KV representation. Instead of storing full per-head keys and values for every token in the cache, the model stores a compact latent vector for each token and reconstructs head-specific keys and values on the fly during attention computation. This shifts long-context efficiency from a token-reduction problem outside the model to a state-compression problem inside attention itself. The full multimodal token stream is preserved, but the memory footprint of cached states is substantially reduced.

\subsubsection{Pretrained Attention Reparameterization}

Our model is built by converting a pretrained GQA-based backbone into the M$^2$LA form. The goal of this conversion is to preserve the original short-context behavior as much as possible while enabling a lower-memory long-context regime.

\paragraph{RoPE-Aware Positional Aggregation.}
For channels carrying positions, compressing keys across heads can distort the geometric structure induced by RoPE~\citep{su2024roformer}. We therefore aggregate multi-head positional keys into a shared representation in an MQA-like manner, using a learned position-compatible linear aggregation obtained from a lightweight calibration pass. This preserves positional structure while avoiding redundant caching of similar positional channels across heads.

\paragraph{Low-Rank Latent Factorization for Content Channels.}
For non-positional content channels, we factorize keys and values through a low-rank bottleneck. Specifically, keys and values are first projected into a compact latent space, which is cached, and then head- or group-specific keys and values are reconstructed via learned up-projections. This decouples the KV-cache size from the original number of heads and head dimensions, allowing us to shrink the cache without discarding multimodal tokens.

\paragraph{Modality-Aware Latent Adaptation.}
A shared latent space is used for both text and vision tokens, but their distributions differ substantially. To accommodate this mismatch, M$^2$LA includes lightweight modality-aware adapters, such as separate affine or linear mappings for text and vision tokens, applied on the cached latent vectors prior to reconstruction. This preserves a unified attention interface while allowing the latent representation to adapt to modality-specific statistics.

\subsubsection{Dynamic Latent Budgeting}

The importance of high-rank visual representations is not uniform across layers or modalities. Early layers often require richer visual detail for alignment and retrieval, while deeper layers rely more heavily on distilled semantic representations. M$^2$LA therefore supports \emph{layer-wise} and \emph{modality-wise} latent dimensions. In practice, we reduce the latent rank more aggressively in layers or modalities where reconstruction fidelity is less critical. This yields a controllable memory--accuracy trade-off and is particularly useful for multimodal agentic rollouts whose context grows over time.

\subsubsection{Compatibility with Head-wise QK-Norm}

Recent backbones such as Qwen3~\citep{yang2025qwen3} employ head-wise QK-Norm~\citep{henry2020query}, applying RMSNorm~\citep{zhang2019root} independently to the query and key vectors of each attention head before RoPE. While beneficial for training stability, this normalization complicates post-training attention reparameterization because it introduces \emph{head-specific} and \emph{input-dependent} scaling factors, making a shared linear aggregation across heads appear invalid.

\paragraph{Conflict with Shared Aggregation.}
For head $h$, the normalized key can be written as
\begin{equation}
\mathrm{RMSNorm}(\mathbf{x}_h) = \frac{\mathbf{x}_h}{\sigma(\mathbf{x}_h)},
\end{equation}
where $\sigma(\mathbf{x}_h)$ is the root-mean-square magnitude of the head input. Since $\sigma(\mathbf{x}_h)$ varies with both the head and the input token, one might expect this to obstruct any static aggregation across heads.

\begin{figure}[t]
    \centering
    \begin{subfigure}[t]{0.48\linewidth}
        \centering
        \includegraphics[width=\linewidth]{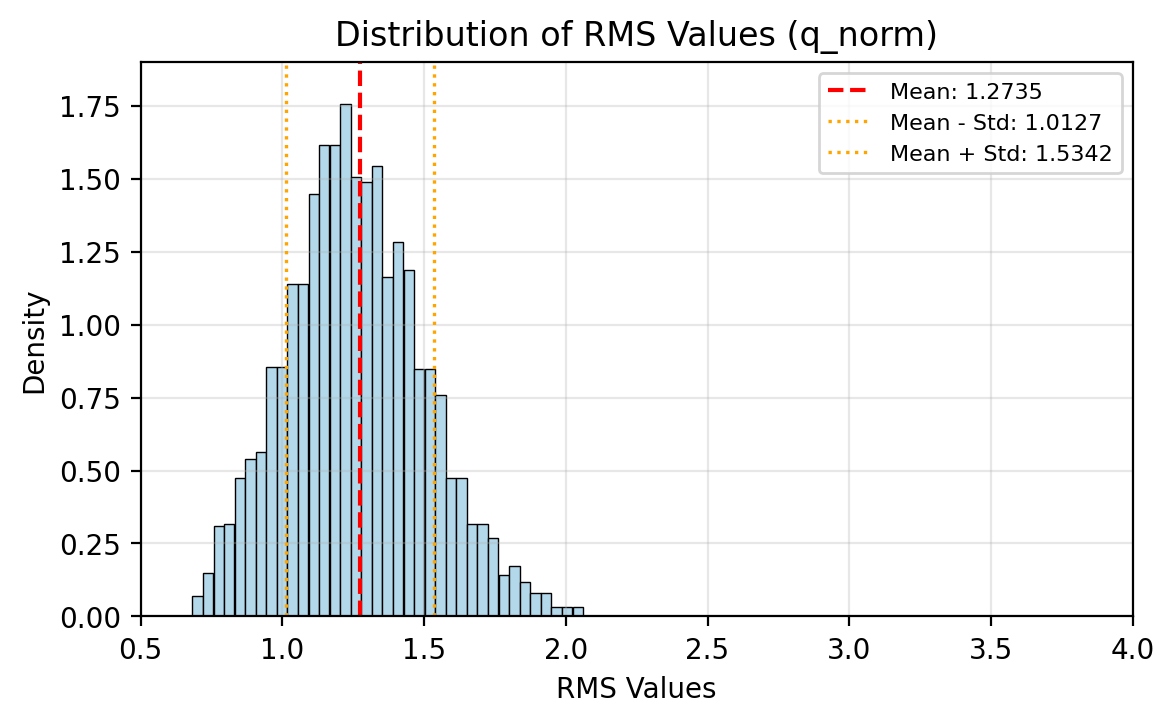}
        \caption{The distribution of RMSNorm coefficients in Qwen3 is highly concentrated.}
        \label{fig:norm_dist}
    \end{subfigure}\hfill
    \begin{subfigure}[t]{0.48\linewidth}
        \centering
        \includegraphics[width=\linewidth]{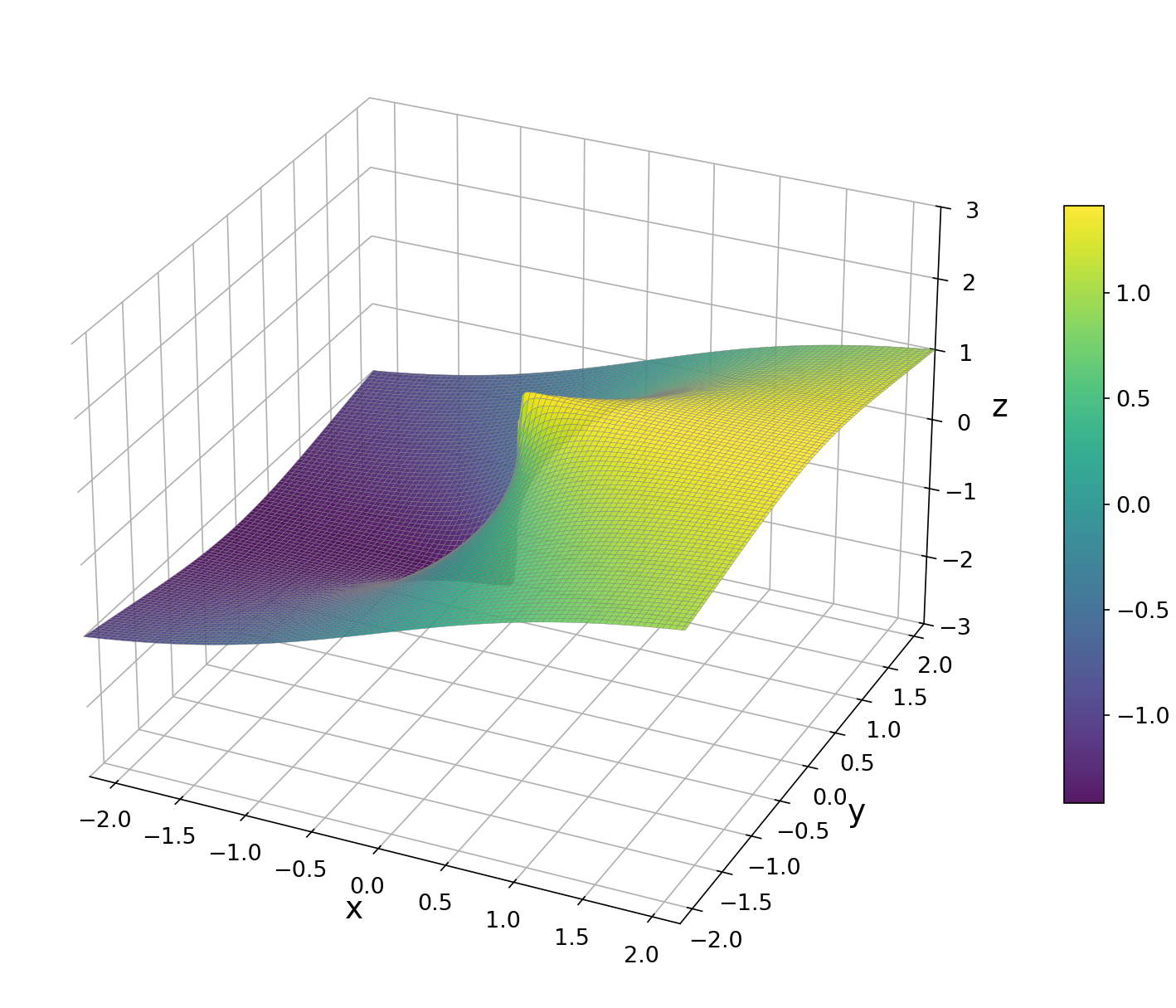}
        \caption{Visualization of 2D RMSNorm $Z = X / \sqrt{(X^2 + Y^2 + \epsilon)/2}$, demonstrating good ``flatness'' and linear approximability.}
        \label{fig:norm_surface}
    \end{subfigure}
    \caption{Empirical validation of RMSNorm concentration and 2D geometric ``flatness''.}
    \label{fig:norm_empirical}
\end{figure}

\paragraph{Norm Concentration and Linear Approximation.}
We address this by observing that in high-dimensional pretrained models, the RMS values are often strongly concentrated. If $\sigma(\mathbf{x}_h)$ has small variance around a mean $\mu$, then a first-order approximation yields
\begin{equation}
\mathrm{RMSNorm}(\mathbf{x}_h)
=
\frac{\mathbf{x}_h}{\sigma(\mathbf{x}_h)}
\approx
\frac{1}{\mu}\mathbf{x}_h.
\end{equation}
Under this approximation, head-wise RMS normalization behaves approximately as a constant linear scaling, which restores compatibility with a shared latent reparameterization.

Empirically, we validate this approximation on a calibration set and find that a learned linear substitute can reproduce the original normalized outputs with high fidelity. This allows us to replace the original head-wise normalization path with a \emph{Global-Norm-Linear} approximation during the M$^2$LA conversion, making Qwen3-style architectures compatible with our attention reparameterization. Overall, M$^2$LA enables substantially longer MCR rollouts by reducing KV-cache memory while keeping the original multimodal token stream intact.

\subsection{Training Recipe for Long-Horizon Multimodal Reasoning}

\begin{figure}[t]
    \centering
    \includegraphics[width=\linewidth]{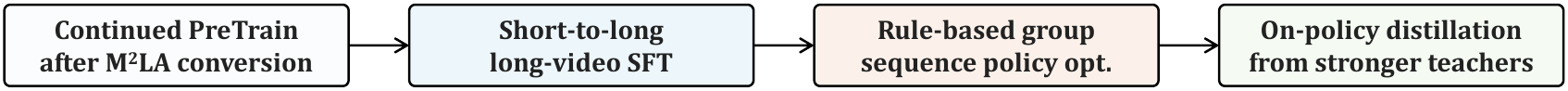}
    \caption{Training recipe of \name.}
    \label{fig:training}
\end{figure}

To realize long-horizon multimodal reasoning after M$^2$LA conversion, we adopt a staged training recipe that progressively restores pretrained capability, extends temporal understanding, and strengthens multi-step reasoning through verifiable post-training.

\subsubsection{Continued Pretraining after M$^2$LA Conversion}

Converting a pretrained GQA backbone into M$^2$LA changes the internal attention parameterization and introduces an initial mismatch relative to the original pretrained model. This mismatch affects both language ability and multimodal alignment, especially before the vision encoder and decoder adapt to the new latent attention pathway. We therefore perform a lightweight \textbf{continued pretraining} (CPT) stage to recover these capabilities.

\subsubsection{Short-to-Long Supervised Fine-Tuning}

After CPT, we conduct a \textbf{short-to-long supervised fine-tuning} stage to progressively build long-range temporal competence. Directly training on maximal-length videos often leads to unstable optimization, since the model must simultaneously learn temporal reasoning and extremely long-context attention. We instead use a curriculum that increases both temporal resolution and context length over stages.

\paragraph{Stage 1: Short Context.}
We begin with videos sampled at 2 fps and capped at 512 frames, corresponding to approximately 32k tokens. This stage establishes basic temporal understanding with manageable computational cost.

\paragraph{Stage 2: Long Context.}
We increase to 4 fps and up to 2048 frames, supporting contexts up to 256k tokens. This stage makes the model reason over extended durations and finer temporal detail.

\subsubsection{Rule-Based Reinforcement Learning}

While SFT builds a stable initialization, it is limited by the scale and diversity of curated annotations. To further improve verifiable long-horizon reasoning, we apply \textbf{rule-based group sequence policy optimization} (GSPO) on automatically rewardable tasks.

\paragraph{Training Tasks.}
We use two families of tasks: 1) \textbf{video QA}, where correctness can be checked against reference answers; and 2) \textbf{temporal grounding}, where predictions can be scored by interval IoU against ground-truth moments. To improve training efficiency, we prioritize examples with meaningful policy uncertainty, filtering out samples that are already solved or too noisy.

\paragraph{Optimization.}
For each video-query pair, we sample a group of candidate responses, compute their rule-based rewards, normalize these rewards within the group, and optimize a clipped policy objective with KL regularization to the SFT reference model. This encourages the policy to improve relative ranking among candidate trajectories while remaining close to the stable supervised initialization. The resulting RL stage strengthens temporal localization, answer calibration, and other aspects of verifiable multimodal reasoning.

\subsubsection{On-Policy Distillation}

Finally, we perform \textbf{video on-policy distillation} from a stronger teacher model. Unlike standard distillation that trains only on teacher-generated outputs, on-policy distillation supervises the student on its own sampled trajectories. For each selected prompt, the student generates a response trajectory and the teacher evaluates the same token sequence under the same prefix. The student is then optimized with a reverse-KL objective toward the teacher distribution:
\begin{equation}
\mathcal{L}_{\text{OPD}} =
\mathbb{E}_{\mathbf{Y}\sim\pi_\theta(\cdot|\mathbf{V},\mathbf{Q})}
\left[
\frac{1}{T}\sum_{t=1}^{T}
\left(
\log \pi_\theta(y_t|\mathbf{V},\mathbf{Q},\mathbf{Y}_{<t})
-
\log \pi_{\text{teacher}}(y_t|\mathbf{V},\mathbf{Q},\mathbf{Y}_{<t})
\right)
\right],
\end{equation}
where $\pi_{\text{teacher}}$ denotes a teacher model with strong capability (here we use Qwen3-235B). 
This provides dense supervision on the states actually visited by the student and reduces exposure bias over long reasoning trajectories.

We construct the distillation set from reasoning-heavy video QA and long-form captioning data, keeping only examples where the teacher is correct or significantly more complete while the student remains incorrect, incomplete, or weakly grounded. This stage is useful for transferring stronger behaviors in multi-step reasoning, fine-grained temporal comparison, and long-form visual description.

\subsection{Video-Agent Instantiation}

To illustrate the practical implications of MCR beyond static QA, we instantiate \name\ as a \textbf{video agent} that alternates between coarse perception, memory retrieval, targeted re-perception, tool use, and answer verification.

\paragraph{Hierarchical Memory.}
Given a video and a query, we first build a hierarchical memory consisting of sampled frames, scene boundaries, clip-level captions, timestamps, and optional subtitle or OCR signals. Each memory entry stores both semantic content and temporal metadata, enabling retrieval by either meaning or time constraints. During rollout, the agent maintains the current evidence window, a retrieved memory subset, and a compact belief summary recording the current hypothesis, unresolved uncertainty, and candidate timestamps.

\paragraph{Question Routing.}
Before acting, the model predicts a coarse question type, such as \texttt{global}, \texttt{speech}, \texttt{knowledge}, \texttt{temporal}, or \texttt{fine-grained}. This serves as a soft routing prior for deciding whether the current memory suffices or whether an external tool should be invoked.

\paragraph{Tool Set.}
The controller may call: 1) \texttt{video\_segmentation}, to update scene-level structure; 2) \texttt{asr}, to retrieve spoken content; 3) \texttt{web\_search}, to obtain external factual knowledge; 4) \texttt{temporal\_grounding}, to localize relevant intervals; 5) and internal actions such as \texttt{summarize}, \texttt{verify}, and \texttt{answer}.
Tool outputs are treated as feedback $\mathbf{F}_{i+1}$ in the MCR framework and appended to the evolving context.

\paragraph{Recursive Verification.}
Before termination, the agent performs a lightweight self-check that asks whether the answer is adequately supported, what evidence backs it, and whether unresolved conflicts remain. If support is insufficient, the agent issues another retrieval or re-perception step focused on the uncertain segment. This verification mechanism reduces hallucination and encourages evidence-grounded responses.

\paragraph{Integration with M$^2$LA.}
The video-agent system is tightly coupled with our long-context architecture. Because M$^2$LA reduces the KV-cache footprint of accumulated evidence, feedback, and belief updates, the agent can retain longer histories without repeatedly rebuilding context from scratch. As a result, simple questions can be answered after a coarse pass, while harder temporal or causal questions can trigger additional recursive evidence gathering within the same long-horizon rollout.

\paragraph{Summary.}
Taken together, MCR provides the \emph{reasoning formulation}, M$^2$LA provides the \emph{efficiency mechanism}, and our staged post-training recipe provides the \emph{capability acquisition path}. This combination yields a more effective long-horizon visual reasoner and supports a practical video-agent instantiation built on top of an open multimodal foundation model.

\section{Data Curation}

We describe below the data used in each stage of the training recipe.

\subsection{Continued Pre-training Data}

The continued pre-training (CPT) corpus comprises \textbf{16M multimodal samples}, corresponding to around \textbf{13.5B tokens}. This stage is not to teach the model new task-specific behaviors, but to \emph{re-stabilize} the converted backbone after the M$^2$LA attention reparameterization and recover its general multimodal capability before downstream long-video training.

We adopt a balanced three-way mixture: 1) text-only data (29.9\% in samples, 3.6B)~\citep{wang2025scireasoner,kuckreja2024geochat,yu2024metamath,hendrycks2020measuring,wei2024measuring}, to recover general language modeling, reasoning, and code ability; 2) image-text pairs (54.9\%, 4.1B)~\citep{Llava-onevision,marti2002iam,yu2016refcoco,hudson2019gqa}, to preserve vision-language alignment and visual grounding; 3) video-caption data (15.2\%, 5.8B)~\citep{zhang2024llava,Maaz2024VideoGPT+,wiedmann2025finevision,yang2025cambrian,wang2023internvid,clark2026molmo2}, to adapt the model to temporal and multimodal inputs under the new attention structure.

\begin{table}[t]
\label{tab:capability}
\resizebox{\textwidth}{!}{%
\begin{tabular}{@{}c l r l r l r l r@{}}
\toprule
Category & \multicolumn{2}{c}{\textbf{Perception \& Recognition}} & \multicolumn{2}{c}{\textbf{Spatial-Temporal Understanding}} & \multicolumn{2}{c}{\textbf{Event \& Action Reasoning}} & \multicolumn{2}{c}{\textbf{Holistic Semantics}} \\
\midrule
\multirow{6}{*}{Sub-category}
& Action Recognition           & 12.5 & Temporal Reasoning          & 11.3 & Action Reasoning        & 8.4 & Plot Understanding   & 7.6 \\
& Object Recognition           & 9.7  & Spatial Perception           & 10.8 & Event Chain Reasoning   & 6.9 & Information Synopsis & 5.5 \\
& Attribute Perception         & 6.2  & Multi-hop Temporal Reasoning & 4.3  & Object Reasoning        & 2.7 &     -                 & -\\
& OCR Problems                 & 5.1  & Scene State Tracking         & 3.8  & Anomaly Recognition     & 0.1 &    -                  & -\\
& Counting Problem             & 1.5  & Egocentric Understanding    & 3.2  &             -            &    -  &      -                & -\\
& Needle Detail                & 0.4  &      -                        &   -    &          -               &   -   &    -                  & -\\
\midrule
\multicolumn{1}{c}{\textbf{Total (100\%)}} & & \textbf{35.4} & & \textbf{33.4} & & \textbf{18.1} & & \textbf{13.1} \\
\bottomrule
\end{tabular}%
}
\centering
\caption{Video understanding capability distribution across perception, spatiotemporal understanding, event reasoning, and holistic semantics.}
\end{table}

\begin{figure}[t]
    \centering
    \includegraphics[width=.9\linewidth]{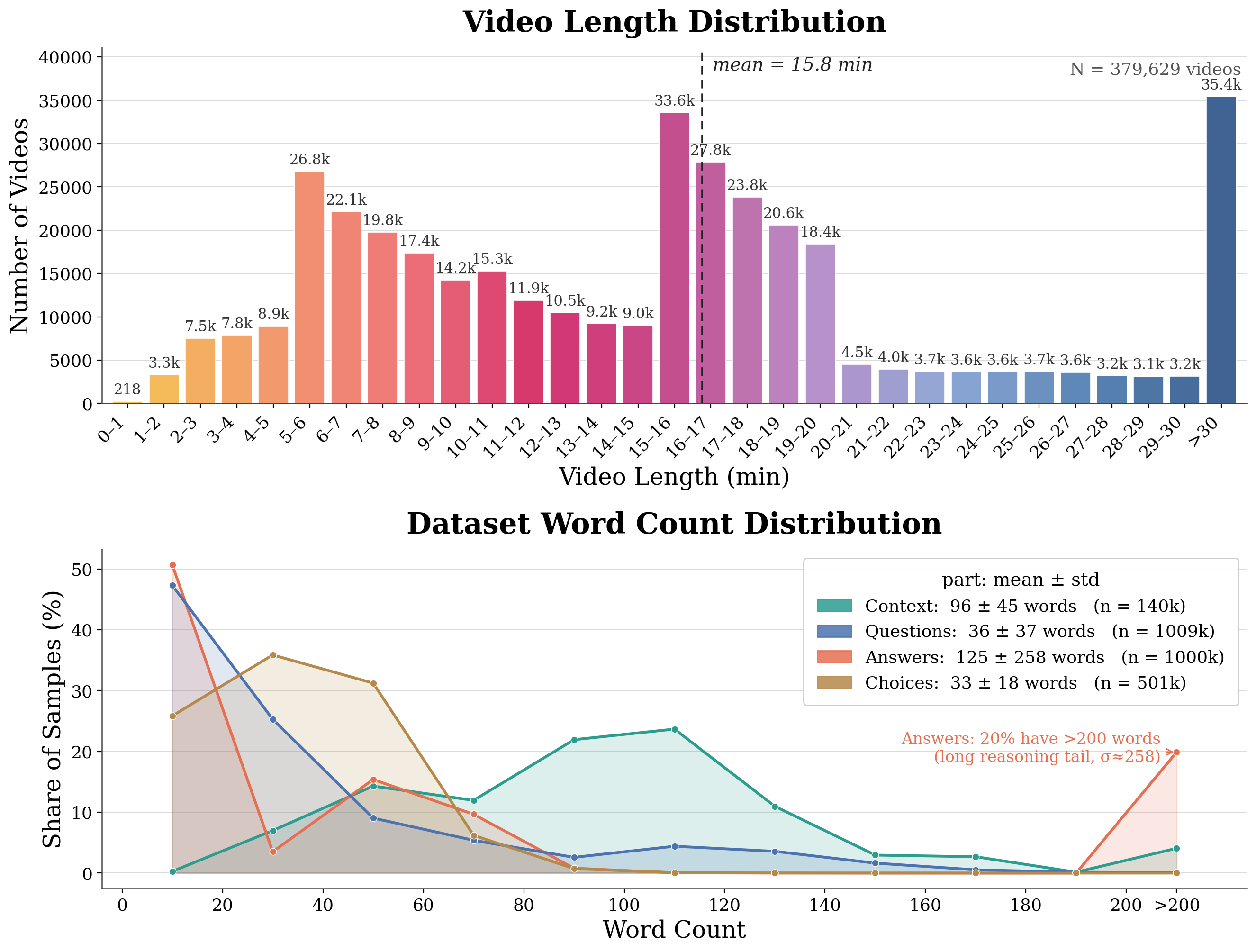}
    \caption{Overview of the long-video supervision corpus. \textbf{Top:} video length distribution ($N=380K$, mean $15.8$ minutes). \textbf{Bottom:} word-count distribution of context, questions, answers, and multiple-choice options.}
    \label{fig:dataset_overview}
\end{figure}

\subsection{Supervised Fine-Tuning Data}

Our supervised fine-tuning (SFT) corpus comprises $\sim$\textbf{7.2M} multimodal samples. The corpus spans long-video understanding~\citep{zhang2024llava}, short-video QA~\citep{caba2015activitynet}, image understanding~\citep{mscoco,lei2018tvqa}, STEM reasoning~\citep{kuckreja2024geochat}, code generation~\citep{gui2025webcode2m}, document comprehension~\citep{mathew2021docvqa,masry2023unichart}, UI grounding~\citep{cheng2024seeclick}, general conversation~\citep{zhang2024llava}, translation~\citep{penedo2026finetranslations}, and spatial reasoning~\citep{yang2025cambrian}. Specifically, the largest components include STEM/science/math (\textbf{22.4\%}), long-video QA (\textbf{17.2\%}), general/conversational data (\textbf{9.9\%}), other video understanding (\textbf{9.1\%}), and image understanding (\textbf{9.1\%}), with additional categories covering translation and language, code generation, chart/document/OCR tasks, UI grounding and long-document understanding, general vision instruction tuning, video reasoning, temporal understanding, and 3D/spatial multimodal tasks. Unless otherwise noted, these datasets are standard public resources or previously used instruction-tuning corpora. Our main data contribution in this stage is the \textbf{long-video supervision component}, which is specifically designed to support long-horizon multimodal reasoning while remaining compatible with broader multimodal and agentic workloads.

\subsubsection{Long-Video Supervised Fine-Tuning Data}

To support long-horizon multimodal reasoning, we curate a large-scale long-video supervised fine-tuning (SFT) corpus through a hierarchical annotation pipeline designed to preserve both \emph{local visual detail} and \emph{global temporal coherence}. The resulting dataset contains \textbf{379K videos} with a mean duration of \textbf{15.8 minutes} (totally $\sim$100K hours), making it better suited to long-context multimodal reasoning than conventional short-video instruction data.

\paragraph{Data Sources.}
The corpus is constructed from complementary long-form video sources spanning both general and reasoning-intensive domains. The largest portion comes from \textbf{InternVid} (\textbf{188K} videos) \citep{wang2023internvid}, covering broad real-world content with video durations typically ranging from 5 minutes to over 30 minutes. We further include \textbf{115K} YouTube-based reasoning videos, which contain rich instructional, explanatory, and cognitively demanding content, and \textbf{77K} specialized videos from V-MME-style~\citep{videomme} tasks, which provide additional supervision for video-centric reasoning and evaluation-aligned capabilities. This mixture gives the training set both broad coverage and strong emphasis on long-form reasoning.

\paragraph{Hierarchical Narrative Annotation.}
Because raw long videos often exceed the reliable context window of current MLLMs, we decompose them into scene-consistent clips using scene-aware temporal splitting. This avoids the semantic fragmentation caused by naive uniform chunking and yields shorter segments that remain temporally coherent. Each clip is captioned by a strong teacher model to produce fine-grained, localized descriptions of actions, objects, scene changes, and salient events.

Clip-level captions provide local evidence but do not yield a coherent global understanding of the video. We therefore perform hierarchical caption merging: neighboring clip captions are first aggregated into scene-level summaries, and these scene summaries are then merged into a long-form narrative that preserves temporal order, cross-scene entity consistency, and high-level event structure. The resulting long-form narratives are compact but globally coherent, with an average narrative context length of 96 words. This hierarchical design provides both \emph{locally grounded supervision} through clip-level captioning and \emph{globally coherent supervision} through cross-temporal narrative integration.

\paragraph{QA Synthesis.}
On top of these hierarchical narratives, we synthesize over \textbf{1M} question-answer pairs spanning four capability dimensions:
\begin{itemize}[leftmargin=*]
    \item \textbf{Perception \& Recognition} (\textbf{35.4\%}): including action recognition, object recognition, attribute perception, OCR, and counting;
    \item \textbf{Spatial-Temporal Understanding} (\textbf{33.4\%}): including temporal reasoning, spatial reasoning, multi-hop temporal inference, scene-state tracking, and egocentric understanding;
    \item \textbf{Event \& Action Reasoning} (\textbf{18.1\%}): including action reasoning, event-chain reasoning, object-centric reasoning, and anomaly detection;
    \item \textbf{Holistic Semantics} (\textbf{13.1\%}): including plot understanding and long-form information synopsis.
\end{itemize}

\paragraph{Analysis.}
This SFT corpus is tailored to the needs of Multimodal Contextual Reasoning. In particular, the annotation pipeline encourages the model to learn from both \emph{fine-grained local evidence} and \emph{cross-temporal global structure}, which is essential for long-horizon multimodal reasoning. The diversity of QA types further encourages the model not only to recognize content, but also to track state over time, connect distant events, and form coherent high-level interpretations of long videos.

\subsection{Post-Training Data}

The post-training stage uses two types of data, corresponding to the two objectives in our recipe: \textbf{rule-based reinforcement learning} and \textbf{on-policy distillation (OPD)}. In both cases, the data is selected to emphasize \emph{informative} long-horizon video examples rather than easy or low-signal samples.

\paragraph{Data for Rule-Based Reinforcement Learning.}
For video RL, we construct training data from two complementary sources: temporal grounding and multiple-choice video QA. These tasks provide automatically verifiable rewards, making them suitable for stable large-scale post-training.

For temporal grounding, we run the SFT model on the TimeLens~\citep{zhang2026timelens} training split and compute the intersection-over-union (IoU) between each predicted temporal span and the ground-truth interval. We retain examples with IoU in the range $[0.1, 0.7]$. This removes both near-trivial cases that the model already solves reliably and severely mismatched cases that are too noisy to provide useful reward signals. From the filtered pool, we sample \textbf{5K} examples for RL training.

For multiple-choice video QA, we sample responses from our SFT multiple-choice data using temperature-$1$ decoding. We retain questions for which the sampled responses contain both correct and incorrect answers, yielding examples with meaningful policy uncertainty and non-degenerate reward structure. This produces a training set of \textbf{10K} questions with verifiable correctness signals.

\paragraph{Data for On-Policy Distillation.}
We focus on examples that reveal a clear teacher--student capability gap. Concretely, we retain samples where the teacher model produces a correct, complete, or well-grounded response, while the student produces an incorrect, incomplete, or weakly grounded one. This filtering ensures distillation is spent on examples that are genuinely informative for the student, rather than on cases the student already handles well or cases where the teacher is unreliable.

This OPD set is drawn from reasoning-intensive video QA and long-form video description data, with an emphasis on examples requiring temporal localization, cross-event comparison, multi-step inference, or dense long-video comprehension. Thus, OPD complements rule-based RL as RL improves performance on verifiable temporal and QA tasks, while OPD transfers stronger teacher behavior on more complex trajectories that are difficult to optimize using hand-designed rewards alone.

\section{Experiments}
We evaluate \name\ from three complementary perspectives. First, we measure \textbf{general multimodal capability} on standard short-video, long-video, and spatiotemporal reasoning benchmarks. Second, we study \textbf{agentic multimodal reasoning} through a video-agent setup with retrieval, grounding, and verification tools. Third, we analyze the contribution of each component in our recipe, including M$^2$LA, short-to-long training, rule-based RL, and on-policy distillation. Across these experiments, our goal is not only to assess static video QA accuracy, but also to test whether the proposed \emph{Multimodal Contextual Reasoning} framework improves the model's ability to reason over evolving visual evidence under long-horizon constraints.

\paragraph{Model.}
Our model is built on a Qwen3-based~\citep{yang2025qwen3} multimodal backbone in the 7/8B parameter regime and converted to the proposed M$^2$LA attention form. We then apply the full staged recipe described in Section~\ref{sec:method}: continued pretraining after M$^2$LA conversion, short-to-long SFT on large-scale video data, rule-based RL, and on-policy distillation from stronger teacher models.

\paragraph{Evaluation Protocol.}
Unless otherwise stated, we follow the official evaluation protocol of each benchmark. For standard benchmark comparisons, we report the scores in their official metric, such as accuracy or average score, and compare against representative open-source video MLLMs. For agentic evaluation, we report task success or answer accuracy under a fixed interaction budget and tool set. Additional implementation details are provided in the appendix.

\paragraph{Benchmarks.}
We evaluate three groups of video-centric tasks, covering long-context understanding, short-video reasoning, and spatiotemporal intelligence:
\begin{itemize}[leftmargin=*]
\item \textbf{Long-video understanding}: We evaluate on Video-MME~\citep{videomme} and VideoMME-v2~\citep{fu2026video}, which provide comprehensive video QA tasks over diverse video durations and domains, emphasizing robust multimodal reasoning. LongVideoBench~\citep{longvideobench}, MLVU~\citep{mlvu}, and LVBench~\citep{wang2024lvbench} further stress long-context modeling, requiring models to retrieve sparse evidence and reason over extended videos. VRBench~\citep{yu2025vrbench} focuses on multi-step reasoning over long narrative videos, while EgoSchema~\citep{egoschema} evaluates long-horizon understanding of egocentric human activities.

\item \textbf{Short-video understanding}: We evaluate on NExT-QA~\citep{xiao2021next} and Perception Test~\citep{perceptiontest}, which test causal, semantic, and multimodal reasoning in short videos. MVBench~\citep{mvbench}, TOMATO~\citep{shangguan2025tomato}, MotionBench~\citep{hong2025motionbench}, and TempCompass~\citep{liu2024tempcompass} focus more explicitly on temporal perception, including event ordering, frame-order sensitivity, motion dynamics, and fine-grained temporal changes.

\item \textbf{Spatiotemporal intelligence}: For temporal grounding, we evaluate on QVHighlights~\citep{qvhighlight}, Charades-STA~\citep{charades-sta}, and ActivityNet Captions~\citep{caba2015activitynet}, which require localizing query-relevant moments or densely grounding events in untrimmed videos. For spatial intelligence, we use VSIBench~\citep{yang2025thinking}, MMSIBench~\citep{yang2025mmsi}, MMSIBench-Video~\citep{lin2025mmsi}, and DSIBench~\citep{zhang2025dsi}, which evaluate spatial configuration understanding, multi-view reasoning, dynamic 3D relations, and spatial memory.
\end{itemize}
We follow the official evaluation protocol of each benchmark and report the official metrics.

\subsection{Main Results}
\subsubsection{Long-Video Understanding}

\begin{table*}[t]
    \centering
\begin{adjustbox}{width=\linewidth,center}
\renewcommand{\arraystretch}{1.1}
\setlength{\tabcolsep}{1.5mm}
\begin{tabular}{lccccccc}
\toprule
\textbf{Model} & \textbf{Video-MME} & \textbf{VideoMME-v2} & \textbf{LongVideoBench} & \textbf{MLVU} & \textbf{LVBench} & \textbf{VRBench} & \textbf{EgoSchema} \\
\midrule
\multicolumn{8}{l}{\textbf{\textit{Proprietary Models}}} \\
GPT-5~\citep{gpt5} & 83.3 & 44.7 & 72.6 & 77.7 & 65.2 & - & 75.6 \\
GPT-5 mini~\citep{gpt5} & 77.3 & - & 69.7 & 69.1 & 54.7 & - & 70.9 \\
Gemini 3 Pro~\citep{gemini3pro} & 88.6 & 56.8 & 75.9 & 75.7 & 77.0 & 76.1 & 68.9 \\
Gemini 2.5 Pro~\citep{comanici2025gemini} & 87.8 & - & 76.8 & 81.5 & 75.7 & - & 72.2 \\
Gemini 2.5 Flash~\citep{comanici2025gemini} & 84.2 & - & 73.1 & 75.1 & 64.9 & - & 70.2 \\
Claude Sonnet 4.5~\citep{sonnet45} & 74.2 & - & 65.1 & 64.0 & 50.5 & - & 73.1 \\
\midrule
\multicolumn{8}{l}{\textbf{\textit{Open-Weight Models}}} \\
VideoChat-Flash-7B~\citep{li2024videochat} & 65.3 & 27.5 & 64.7 & 56.0 & 48.2 & 50.8 & 51.3 \\
InternVL3.5-8B~\citep{internvl35} & 66.0 & 26.0 & 62.1 & 53.2 & 43.4 & 64.1 & 58.6 \\
Qwen3-VL-8B~\citep{qwen3vl} & 71.4 & \textbf{27.9} & 62.4 & 57.6 & \textbf{58.0} & 59.4 & 69.8 \\
Keye-VL-1.5-8B~\citep{keyevl15} & \underline{73.0} & 23.8 & 66.0 & 53.8 & 42.8 & 53.2 & 56.3 \\
GLM-4.1V-9B~\citep{hong2025glm} & 68.2 & 26.9 & 65.7 & 56.6 & 44.0 & 59.8 & 62.6 \\
LLaVA-Video-7B~\citep{zhang2024llava} & 63.3 & 19.9 & 58.2 & 52.8 & 44.2 & - & 57.3 \\
Molmo2-8B~\citep{clark2026molmo2} & 69.9 & 27.4 & \underline{67.5} & 60.2 & 52.8 & - & 62.0 \\
MiniCPM-V-4.5-8B~\citep{yu2025minicpm} & 67.9 & 26.5 & 63.9 & 60.6 & 50.4 & 55.9 & 49.6 \\
Eagle2.5-8B~\citep{chen2026eagle} & 72.4 & 24.9 & 66.4 & 60.4 & 50.9 & 55.7 & \underline{72.2} \\
InternVideo2.5-7B~\citep{wang2025internvideo2} & 65.1 & 26.4 & 60.6 & \underline{72.8} & 46.4 & 51.9 & 63.9 \\
InternVideo3 & \textbf{73.8} & \underline{27.6} & 66.8 & \textbf{77.3} & \underline{55.7} & \textbf{69.4} & \textbf{76.6} \\
\bottomrule
\end{tabular}
\end{adjustbox}
\caption{\textbf{Long-video benchmark results} across long video understanding, captioning, and counting benchmarks. Some entries are unavailable because the corresponding models are not applicable or not evaluated on those benchmarks. The best-performing open-weight model is in \textbf{bold}, and the second best is \underline{underlined}.}
\label{tab:long_video_benchmark_results}
\end{table*}

Table~\ref{tab:long_video_benchmark_results} summarizes results on long-video benchmarks. Overall, \name\ performs strongly across this suite and is particularly competitive on benchmarks that require sustained temporal reasoning, long-range evidence integration, and coherent understanding over extended visual context. Among open-weight models, it achieves the best results on \textbf{Video-MME}, \textbf{MLVU}, \textbf{VRBench}, and \textbf{EgoSchema}, while remaining close to the top tier on \textbf{VideoMME-v2} and \textbf{LongVideoBench}. These evaluations are still standard QA-style benchmarks rather than explicit agentic rollouts, thus they test the underlying long-horizon reasoning abilities that a multimodal agent would need.

Specifically, \name\ achieves the best open-weight score on Video-MME with 73.8, slightly surpassing Keye-VL-1.5-8B (73.0), Eagle2.5-8B (72.4), and its base family Qwen3-VL-8B (71.4). On VideoMME-v2, \name\ obtains 27.6, which is slightly below the strongest open baseline Qwen3-VL-8B (27.9) but still among the top open results. This suggests that our post-training and long-horizon adaptation improve performance substantially on most long-video tasks, while some benchmark-specific strengths of the base model remain difficult to fully preserve or exceed.

On LongVideoBench, \name\ reaches 66.8, ranking among the strongest open 8B-scale models and trailing Molmo2-8B (67.5) by only 0.7 points. On MLVU, \name\ achieves 77.3, outperforming all listed open-weight baselines, including InternVideo2.5-7B (72.8), MiniCPM-V-4.5-8B (60.6), Molmo2-8B (60.2), and Eagle2.5-8B (60.4). This is one of the most notable gains and indicates strong improvement on long-form video reasoning beyond short-range event recognition.

On LVBench, \name\ scores 55.7. This is competitive and above many open baselines, including InternVideo2.5-7B (46.4), but remains below Qwen3-VL-8B (58.0). Since \name\ is built on Qwen3-VL-8B, this result also serves as a useful reminder that our long-horizon adaptation is not uniformly dominant on every benchmark: in some cases, the base model retains dataset-specific advantages that our recipe does not fully improve upon. Even so, the model still achieves a strong overall long-video profile across the broader suite.

Two particularly notable results are \textbf{VRBench} and \textbf{EgoSchema}. On \textbf{VRBench}, \name\ achieves 69.4, the best result among open-weight models and substantially above Qwen3-VL-8B (59.4), InternVL3.5-8B (64.1), and Eagle2.5-8B (55.7). On \textbf{EgoSchema}, \name\ reaches 76.6, the best open-weight result in the table, exceeding Eagle2.5-8B (72.2), Qwen3-VL-8B (69.8), and even the reported human score of 76 on this benchmark. We note that surpassing human performance on a single benchmark should be interpreted cautiously, but it nonetheless indicates that an 8B-scale open model can be highly competitive even against much larger systems in some long-video settings.

Relative to the previous InternVideo2.5-7B baseline, the gains are substantial on several representative long-horizon tasks: \textbf{+8.7} on Video-MME (73.8 vs.\ 65.1), \textbf{+6.2} on LongVideoBench (66.8 vs.\ 60.6), \textbf{+4.5} on MLVU (77.3 vs.\ 72.8), \textbf{+9.3} on LVBench (55.7 vs.\ 46.4), \textbf{+17.5} on VRBench (69.4 vs.\ 51.9), and \textbf{+12.7} on EgoSchema (76.6 vs.\ 63.9). Taken together, these results show that our method is effective in the long-video regime that most directly motivates the paper. While it does not outperform every model on every benchmark, it consistently strengthens long-horizon video understanding and brings an 8B-scale open model close to, and in a few cases beyond, the performance of much larger proprietary systems on selected tasks.

\subsubsection{Short-Video Understanding}

\begin{table*}[t]
    \centering
\begin{adjustbox}{width=\linewidth,center}
\renewcommand{\arraystretch}{1.1}
\setlength{\tabcolsep}{1.5mm}
\begin{tabular}{lccccccc}
\toprule
\textbf{Model} & \textbf{NextQA} & \textbf{PerceptionTest} & \textbf{MVBench} & \textbf{Tomato} & \textbf{MotionBench} & \textbf{TempCompass} & \textbf{Short QA Avg.} \\
\midrule
\multicolumn{8}{l}{\textbf{\textit{Proprietary Models}}} \\
GPT-5~\citep{gpt5} & 86.3 & 79.4 & 74.1 & 53.0 & 65.4 & 80.4 & 73.1\\
GPT-5 mini~\citep{gpt5} & 83.2 & 72.0 & 66.5 & 44.1 & 59.9 & 74.9 & 66.8\\
Gemini 3 Pro~\citep{gemini3pro} & 84.3 & 77.6 & 70.4 & 48.3 & 62.6 & 82.8 & 71.0\\
Gemini 2.5 Pro~\citep{comanici2025gemini} & 85.3 & 78.4 & 70.6 & 48.6 & 62.0 & 81.9 & 71.1\\
Gemini 2.5 Flash~\citep{comanici2025gemini} & 81.8 & 74.7 & 67.0 & 39.1 & 59.3 & 80.2 & 67.0\\
Claude Sonnet 4.5~\citep{sonnet45} & 79.2 & 64.3 & 62.1 & 39.6 & 58.5 & 72.8 & 62.8\\
\midrule
\multicolumn{8}{l}{\textbf{\textit{Open-Weight Models}}} \\
VideoChat-Flash-7B~\citep{li2024videochat} & \textbf{85.5} & 76.5 & 74.0 & 32.5 & \underline{60.6} & 69.4 & 66.4\\

InternVL3.5-8B~\citep{internvl35} & 81.7 & 72.7 & 72.1 & 24.6 & 56.6 & 70.3 & 63.0\\
Qwen3-VL-8B~\citep{qwen3vl} & 83.4 & 72.7 & 68.7 & 35.7 & 56.9 & 74.3 & 65.3\\
Keye-VL-1.5-8B~\citep{keyevl15} & 75.8 & 64.2 & 56.9 & 33.0 & 55.1 & \textbf{75.5} & 60.1\\
GLM-4.1V-9B~\citep{hong2025glm} & 81.3 & 74.2 & 68.4 & 30.0 & 59.0 & 72.3 & 64.2\\
LLaVA-Video-7B~\citep{zhang2024llava} & 83.2 & 68.8 & 58.6 & 24.9 & 54.2 & 66.6 & 59.4\\
MiniCPM-V-4.5-8B~\citep{yu2025minicpm} & 78.8 & 70.9 & 60.5 & 29.8 & 59.7 & 72.7 & 62.1\\
Eagle2.5-8B~\citep{chen2026eagle} & 85.0 & 81.0 & 74.8 & 31.0 & 55.7 & \underline{74.4} & \underline{67.0}\\
InternVideo2.5-7B~\citep{wang2025internvideo2} & 84.9 & 74.9 & \textbf{75.7} & 32.9 & \textbf{60.8} & 70.1 & 66.5\\
InternVideo3 & \textbf{85.5} & \textbf{81.4} & \underline{75.0} & \textbf{37.4} & \underline{60.6} & 74.0 & \textbf{69.0} \\
\bottomrule
\end{tabular}
\end{adjustbox}
\caption{\textbf{Short-video benchmark results} across short video question-answering benchmarks. The best-performing open-weight model is in \textbf{bold}, and the second best is \underline{underlined}.}
\label{tab:short_video_benchmark_results}
\end{table*}

Table~\ref{tab:short_video_benchmark_results} reports results on short-video benchmarks. Although \name\ is designed primarily for long-horizon multimodal reasoning, it also performs strongly on standard short-video QA, indicating that the proposed attention conversion and long-video-oriented post-training do not come at the expense of short-form capability. More importantly, the model achieves the best \textbf{Short QA Avg.} among the listed open-weight models, suggesting that improvements aimed at long-horizon reasoning also transfer to general short-video understanding.

At the aggregate level, \name\ obtains a \textbf{Short QA Avg. of 69.0}, outperforming Eagle2.5-8B (67.0), InternVideo2.5-7B (66.5), VideoChat-Flash-7B (66.4), and Qwen3-VL-8B (65.3). This shows although our method is motivated by long-context and long-video reasoning, the resulting model remains broadly competitive rather than narrowly specialized.

Concretely, \name\ matches the best reported open-weight score on NextQA with 85.5, tying VideoChat-Flash-7B and slightly exceeding Eagle2.5-8B (85.0), InternVideo2.5-7B (84.9), and Qwen3-VL-8B (83.4). On PerceptionTest, \name\ achieves 81.4, the best open-weight result in the table, slightly above Eagle2.5-8B (81.0) and well above Qwen3-VL-8B (72.7) and InternVideo2.5-7B (74.9). On MVBench, \name\ obtains 75.0, which is highly competitive and close to the best open results, though still below InternVideo2.5-7B (75.7). On Tomato, \name\ reaches 37.4, the strongest open-weight result listed, improving over Qwen3-VL-8B (35.7), Keye-VL-1.5-8B (33.0), Eagle2.5-8B (31.0), and InternVideo2.5-7B (32.9).

The remaining two benchmarks show a similar pattern of broad competitiveness without universal dominance. On MotionBench, \name\ scores 60.6, tying VideoChat-Flash-7B and narrowly trailing InternVideo2.5-7B (60.8). On TempCompass, \name\ achieves 74.0, which is competitive but below the best open results from Keye-VL-1.5-8B (75.5) and Eagle2.5-8B (74.4). These cases are worth noting because they show that the gains from our long-horizon training recipe are not uniform across all short-video tasks. In particular, some short-form benchmarks may still reflect strengths of the base model or of alternative data mixtures that are not the main focus of our method.

Compared with InternVideo2.5-7B, \name\ improves on nearly all reported short-video tasks: \textbf{+0.6} on NextQA, \textbf{+6.5} on PerceptionTest, \textbf{+4.5} on Tomato, and \textbf{+3.9} on TempCompass, while remaining essentially comparable on MVBench and MotionBench. Relative to its base family Qwen3-VL-8B, the gains are also consistent: \textbf{+2.1} on NextQA, \textbf{+8.7} on PerceptionTest, \textbf{+6.3} on MVBench, \textbf{+1.7} on Tomato, \textbf{+3.7} on MotionBench, and a small \textbf{-0.3} on TempCompass. Overall, these results indicate that the model's improvements are not confined to long-video evaluation alone. Instead, better context handling and post-training for long-horizon reasoning appear to also strengthen short-video perception, temporal reasoning, and answer calibration.

\begin{table*}[t]
    \centering
\begin{adjustbox}{width=\linewidth,center}
\renewcommand{\arraystretch}{1.1}
\setlength{\tabcolsep}{1.5mm}
\begin{tabular}{lcccccc}
\toprule
\textbf{Model} & \multicolumn{3}{c}{\textbf{Temporal Grounding}} & \multicolumn{3}{c}{\textbf{Spatial Intelligence}} \\
\cmidrule(lr){2-4}\cmidrule(lr){5-7}
 & \textbf{QVHighlights} & \textbf{Charades-STA} & \textbf{ANet} & \textbf{VSIBench} & \textbf{MMSIBench} & \textbf{MMSIBench-Video} \\
\midrule
\multicolumn{7}{l}{\textbf{\textit{Proprietary Models}}} \\
GPT-5~\citep{gpt5} & 56.8 & 40.5 & 42.9 & \textemdash & 41.9 & 36.8\\
GPT-5 mini~\citep{gpt5} & \textemdash & \textemdash & \textemdash & \textemdash & 34.9 & \textemdash\\
Gemini 3 Pro~\citep{gemini3pro} & \textemdash & \textemdash & \textemdash & \textemdash & \textemdash & 38.0\\
Gemini 2.5 Pro~\citep{comanici2025gemini} & 70.4 & 52.8 & 58.1 & \textemdash & 37.6 & \textemdash\\
Gemini 2.5 Flash~\citep{comanici2025gemini} & 64.3 & 48.6 & 52.5 & \textemdash & 33.1 & 35.4\\
\midrule
\multicolumn{7}{l}{\textbf{\textit{Open-Weight Models}}} \\
VideoChat-Flash-7B~\citep{li2024videochat} & 32.7 & 39.7 & 24.8 & 32.8 & 27.6 & 28.2 \\
InternVideo2.5-7B~\citep{wang2025internvideo2} & 32.7 & 39.7 & 24.8 & 33.4 & 26.9 & 27.3 \\

InternVL3.5-8B~\citep{internvl35} & 31.3 & 27.8 & 31.3 & 56.0 & \textbf{30.5} & 29.2 \\
Qwen3-VL-8B~\citep{qwen3vl} & 59.4 & 48.3 & 46.8 & 59.1 & 27.0 & 27.6\\
Keye-VL-1.5-8B~\citep{keyevl15} & 55.5 & 45.4 & 41.3 & 36.4 & 26.6 & 26.5 \\
MiniCPM-V-4.5-8B~\citep{yu2025minicpm} & 28.5 & 30.0 & 22.7 & 32.0 & \underline{28.1} & 23.4\\
InternVideo3 & \textbf{59.9} & \textbf{50.4} & \textbf{47.9} & \textbf{68.1} & {27.6} & \textbf{30.7}\\
\bottomrule
\end{tabular}
\end{adjustbox}
\caption{\textbf{Spatiotemporal intelligence results} across temporal grounding and spatial understanding benchmarks.}
\label{tab:spatiotemporal_intelligence_results}
\end{table*}

\subsubsection{Spatiotemporal Intelligence}

Table~\ref{tab:spatiotemporal_intelligence_results} reports results on temporal grounding and spatial intelligence benchmarks. Overall, \name\ shows strong performance on \emph{temporal grounding}, where it achieves the best results among the listed open-weight models on all three benchmarks. These evaluations are particularly relevant to our motivation as they test whether the model can identify \emph{when} and \emph{where} relevant evidence appears, rather than only producing the correct final answer. Although the experiments here are still direct benchmark evaluations rather than full agentic interactions, these localization and grounding abilities are important ingredients for downstream multimodal agents.

On the three \textbf{temporal grounding} benchmarks, \name\ achieves 59.9 on QVHighlights, 50.4 on Charades-STA, and 47.9 on ANet, establishing the best open-weight results in the table on all three tasks. Compared with its base model Qwen3-VL-8B, the gains are consistent though moderate: \textbf{+0.5} on QVHighlights, \textbf{+2.1} on Charades-STA, and \textbf{+1.1} on ANet. Compared with earlier open baselines such as VideoChat-Flash-7B and InternVideo2.5-7B, the improvements are much larger. At the same time, proprietary Gemini 2.5 Pro remains clearly ahead on this suite, reaching 70.4/52.8/58.1. We therefore do not claim to close the gap to the strongest proprietary systems, but the results do show that our approach yields a strong and consistent grounding capability within the open 8B regime.

On the \textbf{spatial intelligence} side, \name\ achieves 68.1 on VSIBench, 27.6 on MMSIBench, and 30.7 on MMSIBench-Video. The strongest result here is on VSIBench, where \name\ outperforms all listed baselines by a clear margin, including Qwen3-VL-8B (59.1) and InternVL3.5-8B (56.0). On MMSIBench-Video, \name\ also obtains the best result among the listed models. On MMSIBench, however, \name\ is competitive but not best, trailing InternVL3.5-8B (30.5) and slightly below MiniCPM-V-4.5-8B (28.1). This again reflects the method improves the model substantially on many spatiotemporal tasks, but does not dominate every individual benchmark.

These results are important beyond standalone benchmark scores. A model can answer video QA questions reasonably well by relying on priors or coarse summaries, yet still fail to localize the supporting evidence in time or space. In contrast, temporal grounding and spatial intelligence more directly measure whether the model can connect answers to the relevant evidence. This distinction matters for multimodal agents: before an agent can decide what tool to call, what segment to revisit, or whether a conclusion is sufficiently supported, it must first know \emph{where} and \emph{when} the evidence lies. From this perspective, the strong grounding and spatial results of \name\ provide useful support for the broader claim that better multimodal contextual reasoning can strengthen capabilities relevant to future agentic behavior.

\subsection{Inference Efficiency}

\begin{figure}[!ht]
    \centering
    \begin{subfigure}[t]{\linewidth}
        \centering
        \includegraphics[width=.65\linewidth]{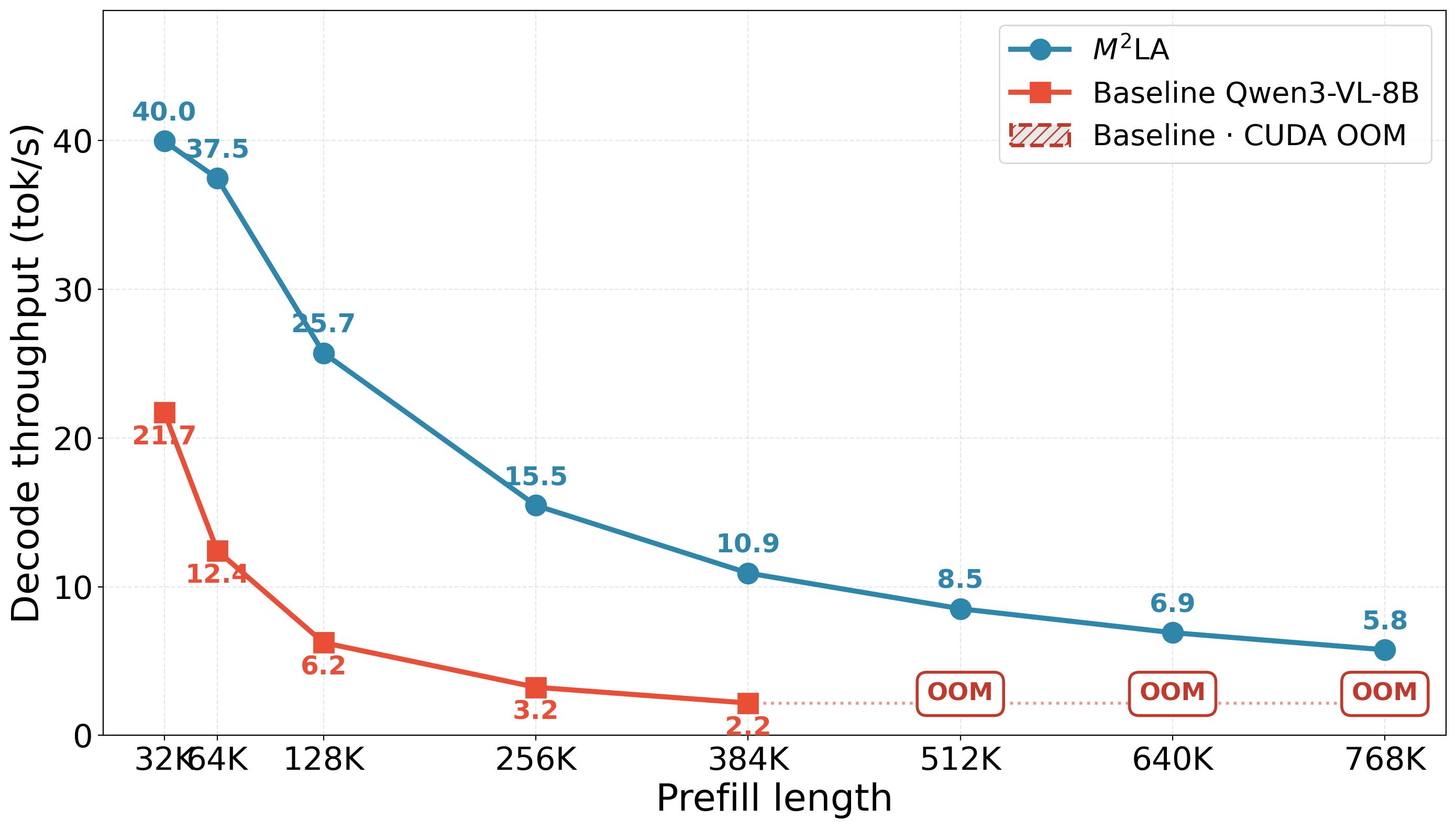}
        \caption{Decode throughput.}
        \label{fig:inference_decode_throughput_a}
    \end{subfigure}

    \vspace{0.5em}
    \begin{subfigure}[t]{\linewidth}
        \centering
        \includegraphics[width=.8\linewidth]{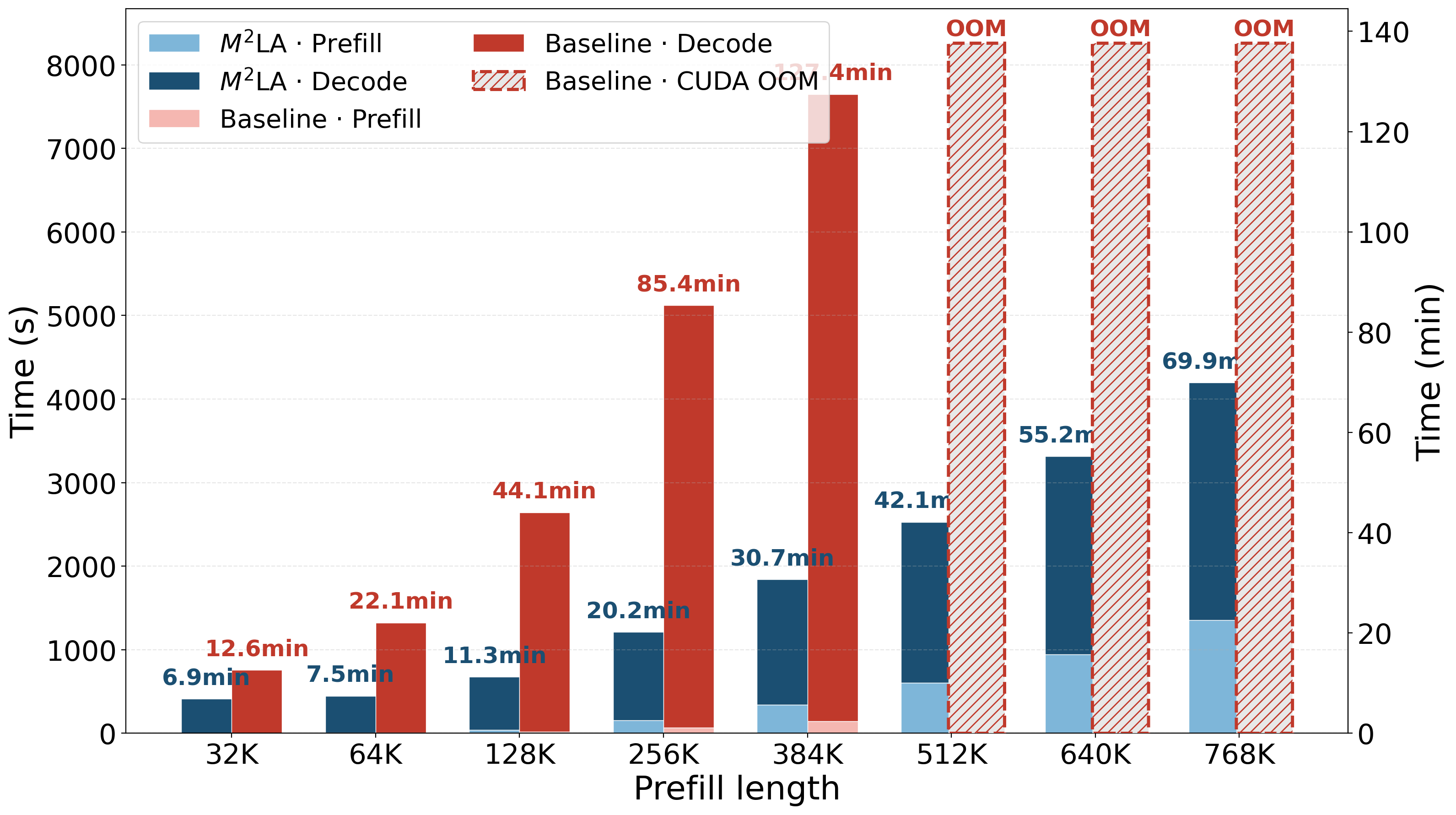}
        \caption{End-to-end latency, decomposed into prefill and decode time.}
        \label{fig:inference_decode_throughput_b}
    \end{subfigure}
    \caption{Inference efficiency of the Qwen3-VL-8B and its M$^2$LA-converted variant on a single H200 GPU. Prefill length varies from 32K to 768K tokens, with decode length fixed to 16K. For 256K+ inputs, both models use 64K chunked prefill and no external inference acceleration. M$^2$LA consistently improves long-context decoding efficiency and remains executable up to 768K prefill tokens, whereas the Qwen3-VL runs out of memory from 512K onward.}
    \label{fig:inference_decode_throughput}
\end{figure}

We evaluate the inference efficiency of M$^2$LA during long-context generation. Specifically, we compare the original Qwen3-8B backbone with its M$^2$LA-converted counterpart under identical inference settings. All measurements are conducted on a single H200 GPU, using one warm-up run followed by ten measured runs. We vary the prefill length from 32K to 512K tokens while fixing the decode length to 16K. No external inference acceleration is applied to either model. For inputs of 256K tokens and above, both models use the same chunked prefill strategy with a chunk size of 64K.

Figure~\ref{fig:inference_decode_throughput} shows that M$^2$LA consistently improves decoding efficiency in the long-context regime. At 32K prefill tokens, the converted model reaches 39.96 tok/s during decoding, compared with 21.74 tok/s for the original model, corresponding to a 1.84$\times$ speedup. As the context grows, the benefit becomes more pronounced: decode throughput improves by 4.12$\times$ at 128K, 4.77$\times$ at 256K, and 5.01$\times$ at 384K prefill tokens. In terms of end-to-end latency, M$^2$LA reduces total runtime by 1.83$\times$ at 32K and by more than 4$\times$ in the 256K--384K regime.

The largest advantage appears at the longest contexts. The original model runs out of memory at 512K prefill tokens, whereas the M$^2$LA remains executable and completes the 16K-token decode phase at 8.52 tok/s. This behavior is consistent with the KV-cache analysis under batch size 1 and bf16 precision, where M$^2$LA reduces the KV-cache footprint by roughly 50\% across context lengths.

These results validate the practical efficiency benefit of compressing long-context KV states. In particular, the gains are most significant in the \emph{decode-dominated} regime that arises in long-horizon multimodal reasoning, where the model must preserve large multimodal contexts while generating long intermediate traces, tool calls, and final responses.

\subsection{Agentic Video Exploration}

In addition to benchmark evaluation, we instantiate \name\ as a video agent using MCR with access to segmentation, ASR, temporal grounding, search, summarization, and verification tools. Although we leave large-scale quantitative agent evaluation to future work, this setup provides a qualitative illustration of the recursive evidence-gathering behavior enabled by our formulation.

\begin{figure}[t]
    \centering
    \includegraphics[width=\linewidth]{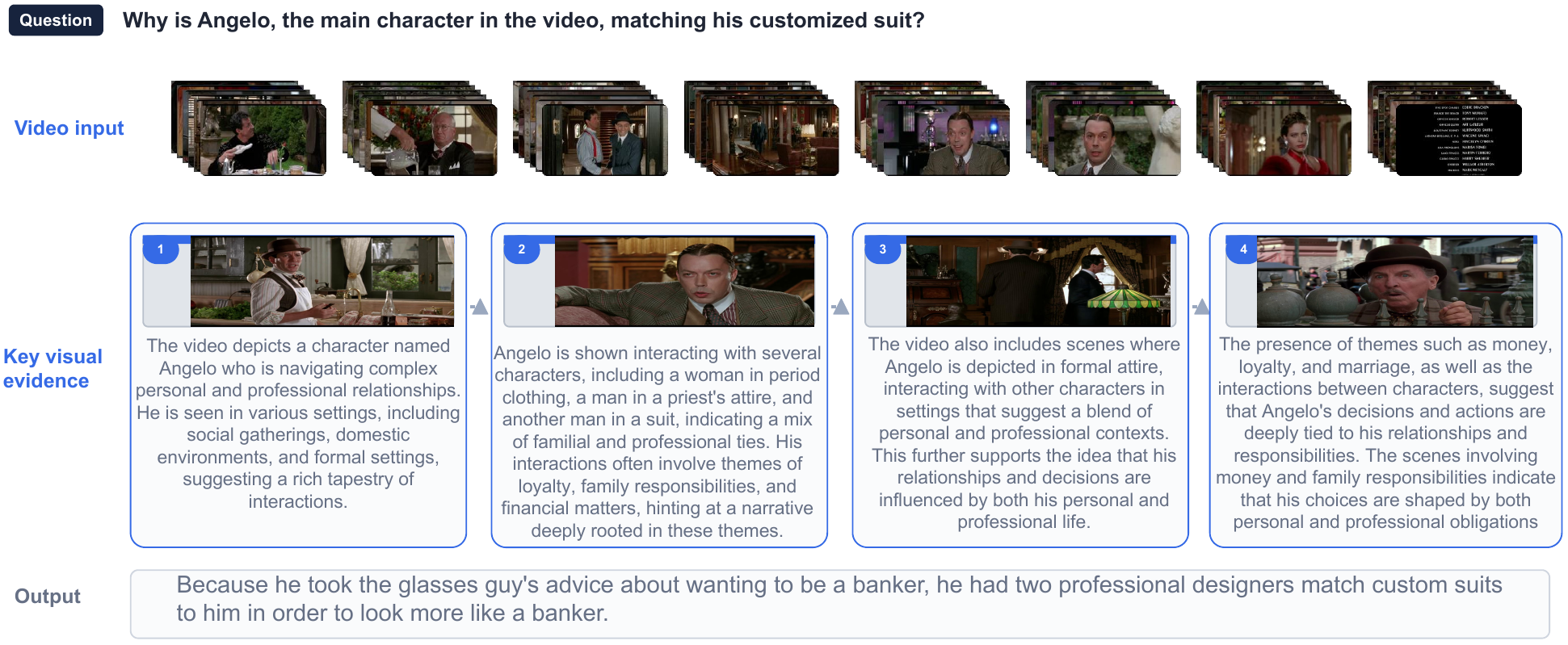}
    \caption{\textbf{Event Attribution.} A qualitative demo on long-video QA: the model is asked why Angelo is fitting a custom suit, and infers that he wants to look more like a banker after following advice from the man with glasses.}
    \label{fig:agentic_demo1}
\end{figure}

\begin{figure}[t] 
    \centering
    \includegraphics[width=\linewidth]{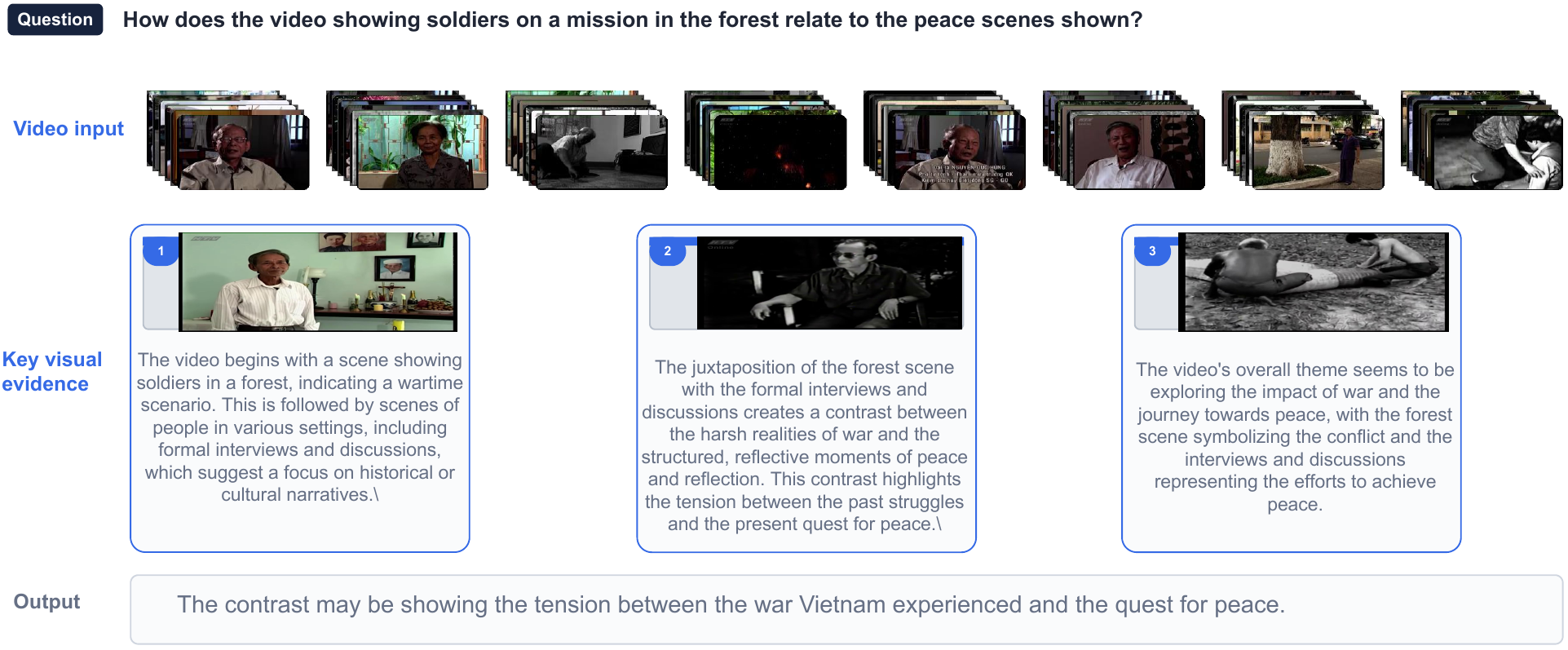}
    \caption{\textbf{Logical Linkage.} A qualitative demo on thematic reasoning: the model is asked how soldiers on a forest mission relate to later peace scenes, and explains that the contrast reflects wartime tension and the pursuit of peace.}
    \label{fig:agentic_demo2}
\end{figure}

\begin{figure}[t]
    \centering
    \includegraphics[width=\linewidth]{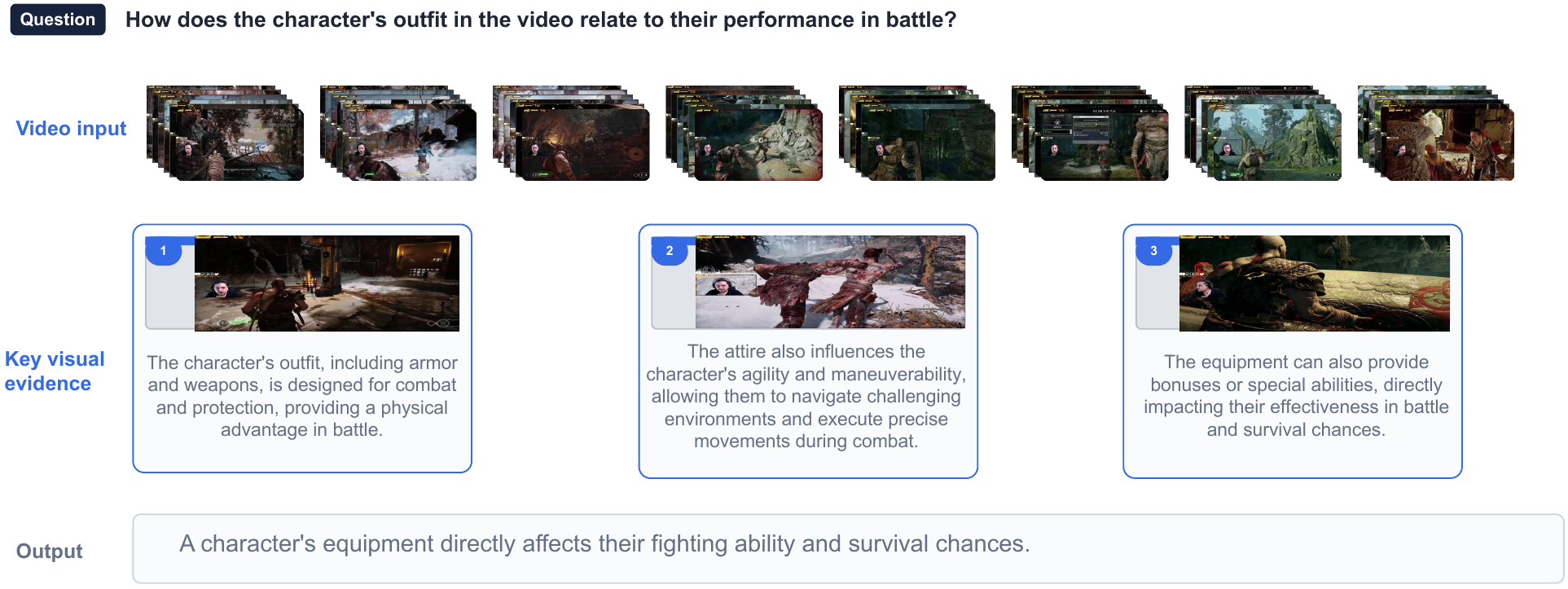}
    \caption{\textbf{Relational Reasoning.} A qualitative demo on battle understanding: the model is asked how a character's outfit affects combat performance, and answers that armor, weapons, and gear directly improve fighting ability and survival chances.}
    \label{fig:agentic_demo3}
\end{figure}

\begin{figure}[t]
    \centering
    \includegraphics[width=\linewidth]{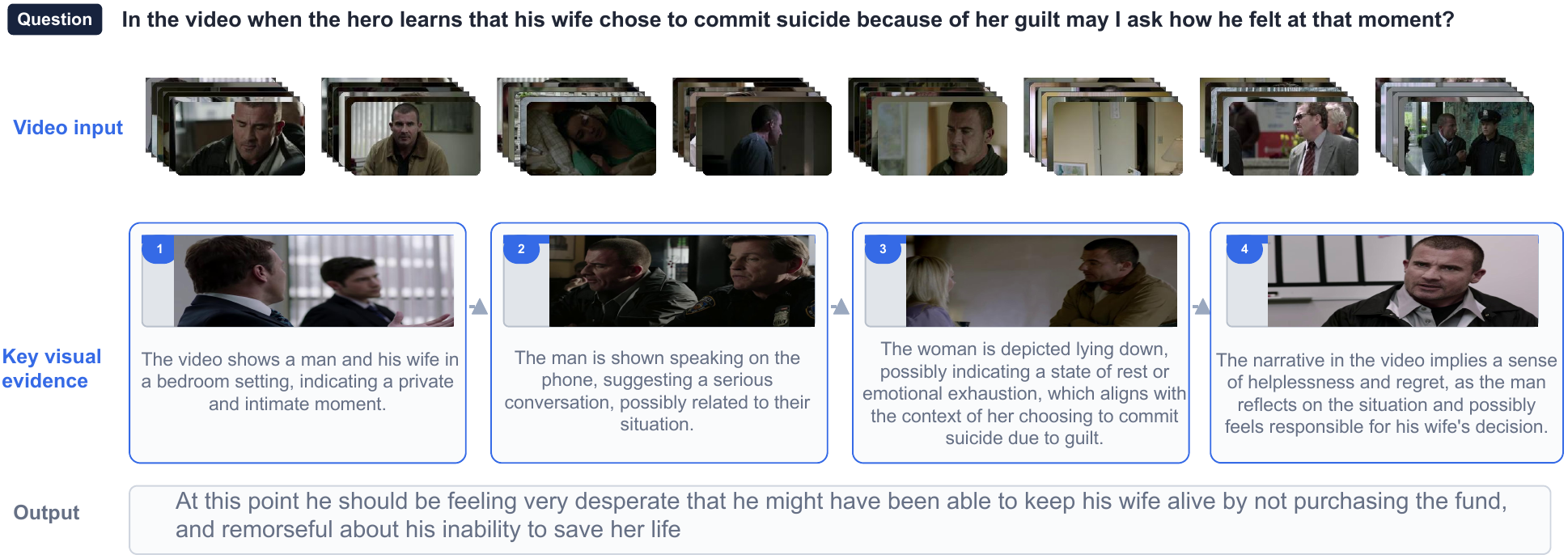}
    \caption{\textbf{Implicit Inference.} A qualitative demo on emotion understanding: the model is asked how the hero feels after learning that his wife chose suicide out of guilt, and answers that he feels desperation and remorse.}
    \label{fig:agentic_demo4}
\end{figure}

Figures~\ref{fig:agentic_demo1}--\ref{fig:agentic_demo4} show four representative agentic video exploration examples. These demos cover complementary forms of long-horizon multimodal reasoning: attributing a character's action to earlier contextual cues, linking temporally distant scenes through a shared theme, relating visual equipment to downstream battle performance, and inferring a protagonist's emotional state from implicit narrative evidence. In each case, the agent grounds its answer in selected visual evidence rather than relying on a single global video summary, illustrating how iterative evidence collection helps resolve questions that require cross-scene context and implicit reasoning.

For long-form video questions, the agent first consults coarse hierarchical memory, identifies uncertain or relevant segments, and then issues targeted tool calls. Speech-heavy questions trigger ASR, event-centric questions trigger temporal grounding, and low-confidence answers trigger a verification step before termination. Compared with a single-pass baseline, this iterative procedure more reliably recovers from incomplete first-pass evidence and reduces errors caused by relying on overly coarse summaries.

These observations support the motivation of MCR: in realistic multimodal settings, the problem is often not only to generate an answer from a fixed prompt, but to determine whether the current evidence is sufficient, what additional evidence is needed, and how that new evidence should update the current belief state.

\subsection{Ablation Studies}

\begin{table}[t]
    \centering
    \renewcommand{\arraystretch}{1.1}
    \setlength{\tabcolsep}{2.2mm}
    \small
    \begin{tabular}{lccccc}
        \toprule
        \textbf{Variant} & \textbf{V-MME} & \textbf{LVB} & \textbf{MLVU} & \textbf{LVBench} & \textbf{Avg.} \\
        \midrule
        Full & \textbf{73.1} & \textbf{64.6} & \textbf{75.4} & 54.2 & \textbf{66.8} \\
        w/o CPT & 68.9 & 62.6 & 72.4 & \textbf{54.7} & 64.7 \\
        w/o Long Ctx. & 70.4 & 61.6 & 75.1 & 54.5 & 65.4 \\
        w/o LV Data & 70.9 & \textbf{64.6} & 72.1 & 51.4 & 64.8 \\
        
        \bottomrule
    \end{tabular}
    \caption{Ablation results on representative long-video benchmarks. V-MME and LVB denote Video-MME and LongVideoBench, respectively. We report the official score for each benchmark and the arithmetic average over the four tasks.}
    \label{tab:ablation_studies}
\end{table}

\begin{table}[t]
    \centering
    \small
    \renewcommand{\arraystretch}{1.1}
    \setlength{\tabcolsep}{4mm}
    \begin{tabular}{lc}
        \toprule
        \textbf{Inference Setting} & \textbf{Video-MME} \\
        \midrule
        Direct QA & 73.1 \\
        Agentic inference (+ MCR) & \textbf{75.8} \\
        \bottomrule
    \end{tabular}
    \caption{Preliminary comparison between direct QA and agentic inference on Video-MME. The agentic setting augments the base model with MCR at inference time.}
    \label{tab:agentic_videomme}
\end{table}

To better understand which parts of our recipe matter most for long-horizon video reasoning, we conduct ablations on three representative ingredients: continued pretraining after M$^2$LA conversion (CPT), long-context training (Long Ctx.), and curated long-video supervision (LV Data). Results on four representative long-video benchmarks are shown in Table~\ref{tab:ablation_studies}.
The full model achieves the best overall average (66.8), indicating that the different stages of our recipe provide complementary gains rather than redundant improvements.
Removing CPT causes the largest drop on Video-MME (-4.2) and a clear degradation on MLVU (-3.0), reducing the overall average from 66.8 to 64.7. This suggests that continued pretraining is an important recovery stage after M$^2$LA conversion: without it, the model does not fully restore its language-side ability and multimodal alignment before downstream long-video specialization. Interestingly, the effect on LVBench is small, which indicates that some benchmark gains can still be obtained from later stages, but the broader long-video capability is noticeably weaker.
Removing long-context training lowers the average to 65.4, with the largest drop on LongVideoBench (-3.0). This is consistent with the purpose of the short-to-long curriculum: the model needs explicit exposure to extended contexts in order to make effective use of the efficient attention structure. In other words, M$^2$LA provides the \emph{capacity} for longer reasoning windows, but long-context training is needed for the model to actually learn how to use them.
Removing curated long-video data reduces the average to 64.8. The largest degradations appear on MLVU (-3.3) and LVBench (-2.8), while LongVideoBench remains unchanged in this setting. This suggests that long-video supervision provides complementary benefits beyond architecture-level context efficiency and context-length scaling: it improves the model's ability to connect distant events, maintain temporally coherent state, and answer questions grounded in long-form narrative structure.

Overall, the ablation results support the design logic of the full recipe. Architectural conversion alone is not sufficient; it must be paired with recovery training, explicit long-context exposure, and broad long-video supervision. These components contribute in different ways, and their combination yields the strongest long-horizon performance.

\paragraph{Agentic Inference.} In addition to the standalone model ablations above, we conduct a small preliminary study of \emph{agentic inference} by using \name\ with MCR at test time. On Video-MME, this increases performance from 73.1 to 75.8 (\textbf{+2.7}). This result is encouraging because it suggests that the model's underlying long-horizon reasoning can be further improved when inference is allowed to revisit evidence and verify intermediate conclusions, consistent with the motivation of MCR.

At the same time, we do not present this as evidence of uniformly better performance across long-video benchmarks. In our current experiments, the agentic setting improves Video-MME but does not consistently help other benchmarks. We therefore view this result as a \emph{proof of concept} rather than a general claim about agentic superiority. One plausible reason is that today's public video benchmarks are still designed primarily for direct question answering and do not always reward multi-step retrieval, verification, or iterative evidence gathering. More systematic evaluation of multimodal agents will likely require benchmarks that explicitly measure such behavior.

Even so, the Video-MME result is useful in two ways. First, it shows that the MCR formulation is not only a conceptual framing for training, but can also be instantiated at inference time in a way that yields measurable gains. Second, it suggests that the direct QA improvements reported in the main benchmark tables are likely conservative with respect to the model's full agentic potential, since current benchmark formats only partially expose the benefits of recursive evidence gathering.

\subsection{Discussion}

Our experiments suggest that longer context alone is not enough. What matters is whether the model can \emph{use} long context as an evolving reasoning substrate. The strongest gains of \name\ appear not simply when more frames are available, but when success depends on maintaining a coherent account of what has happened, what evidence matters, and how observations from distant parts of the video relate to one another.

This is also why our improvements are most substantial on long-video benchmarks and selective temporal reasoning tasks rather than uniformly across all short-form tasks. We view this as evidence for the paper's central claim: long-horizon multimodal agency is a distinct capability axis, and improving it requires more than scaling static multimodal QA.

\section{Conclusive Remarks}

We presented \name, a framework for improving long-horizon multimodal reasoning through \emph{Multimodal Contextual Reasoning} (MCR), efficient long-context attention, and staged post-training. Our central premise is that many video-centric and visually grounded tasks are better modeled not as one-shot multimodal question answering, but as a closed-loop process of evidence accumulation, context updating, tool interaction, and self-correction. Under this view, multimodal understanding becomes a form of recursive contextual reasoning over evolving observations, intermediate conclusions, and external feedback.

To make such rollouts practical, we introduced \emph{Multimodal Multi-head Latent Attention} (M$^2$LA), which compresses KV-cache states while preserving the full multimodal token stream. We further developed a training recipe combining continued pretraining after attention conversion, short-to-long supervised fine-tuning, rule-based reinforcement learning, and on-policy distillation. Together, these components enable a moderate-scale open multimodal backbone to better handle long videos, dense temporal evidence, and extended reasoning trajectories.

Empirically, \name\ achieves strong performance across both short-video and long-video benchmarks, with notable gains on long-horizon tasks such as Video-MME, MLVU, and EgoSchema. Relative to InternVideo2.5-7B, it substantially improves performance on several representative long-video benchmarks, indicating that long-horizon capability can be strengthened through better context handling and long-video-oriented post-training even without moving to a much larger or newer backbone family. We also instantiate the model as a video agent with retrieval and verification tools, illustrating how recursive multimodal reasoning can support more robust evidence-grounded behavior.

More broadly, we view this work as a step \emph{toward} multimodal agents rather than a claim of solving multimodal agency in full generality. Our setting is video-centric, but it highlights a capability that is likely to matter well beyond video: the ability to maintain, update, and verify a multimodal contextual state over long horizons. In this sense, we believe long-video and spatiotemporal reasoning provide a valuable testbed for studying visually grounded agency and world-model-relevant capabilities in foundation models.

At the same time, the open-weight ecosystem has evolved rapidly, and newer model families increasingly adopt stronger native long-context architectures and broader post-training recipes. We therefore view our contribution not as a claim of frontier-wide architectural finality, but as evidence for two broader lessons: first, that \textbf{context efficiency remains crucial} for long-horizon multimodal rollouts; and second, that \textbf{closed-loop multimodal reasoning matters} when models must perceive, remember, verify, and act over extended visual streams. We hope this work helps motivate further research on adapting open multimodal models into more capable and practically useful systems for long-horizon visually grounded tasks.

\section{Limitations, Discussion, and Future Work}

\paragraph{Rapidly Evolving Open-Weight Frontier.}
A primary limitation of this work is that the open-weight multimodal ecosystem evolved rapidly during the course of the project. Newer model families such as Qwen3.5/3.6/3.7~\citep{qwen35blog}, GLM~\citep{zeng2026glm}, Kimi~\citep{team2026kimi}, StepFun~\citep{huang2026step}, and others~\citep{deepseekai2026deepseekv4, mimov25, minimax2026m3, meituanlongcatteam2026longcatnextlexicalizingmodalitiesdiscrete} now surpass our system on a range of benchmarks and often incorporate stronger native long-context designs. Accordingly, this paper should not be read as claiming state-of-the-art multimodal agency or overall frontier open-model performance. Rather, it studies how far a moderate-scale open multimodal model can be pushed toward long-horizon visually grounded reasoning through better contextual reasoning, long-context efficiency, and post-training.

\paragraph{Specific Efficiency Path Versus Broader Efficiency Principle.}
Our long-context route centers on converting a pretrained GQA backbone into a latent-attention form through M$^2$LA. This was a practical choice when the project began, but it is less directly applicable to newer models that already incorporate native MLA~\citep{liu2024deepseek,guo2025deepseek}, linear attention~\citep{team2025kimi}, or more hierarchical compressed-attention designs \citep{deepseekai2026deepseekv4}. Even so, we believe the broader principle remains important: long-horizon multimodal reasoning requires efficient context handling, and there is continued value in studying how pretrained attention can be adapted into more deployment-friendly forms.

\paragraph{Conceptual Novelty of MCR.}
The ideas underlying MCR, including closed-loop reasoning, tool use, memory, self-correction, and iterative evidence gathering, are now widely discussed in agent papers and system overviews~\citep{masterman2024landscape, wang2024survey, zhang2025survey, ning2025survey, ning2026code}. Our contribution is therefore less about inventing these ingredients from scratch and more about offering a clear and extensible formulation for long-horizon multimodal reasoning, especially in video-centric settings. We believe this formulation is useful for training, system design, and evaluation, even if many of its high-level ingredients are increasingly shared across the literature.

\paragraph{Implicit Rather Than Full Predictive World Modeling.}
We position MCR as a form of contextual belief-state modeling relevant to world-model-like capabilities~\citep{vjepa,vjepa2}, not as a full action-conditioned simulator of the environment~\citep{hafner2019dream,hafner2023mastering}. This makes the approach practical for multimodal reasoning tasks, but it also limits its scope relative to predictive world models that explicitly model future environment dynamics. A promising direction is to connect contextual reasoning more tightly with learned predictive models so that multimodal agents can combine evidence gathering with explicit future-state evaluation.

\paragraph{Tool Dependence and System Fragility.}
The agent behavior depends on external tools such as ASR, temporal grounding, segmentation, retrieval, and search. Errors from these tools can propagate into the shared context and distort later reasoning. In the current work, tool interfaces are lightweight and not jointly optimized with the base model. Future work may explore tighter model--tool integration, uncertainty-aware tool use, and better calibration of tool-derived evidence.

\paragraph{Scope of the Current Evaluation.}
Our experiments focus primarily on long-video understanding, short-video reasoning, and a video-agent setup with perception tools. This covers an important but still partial slice of multimodal agency. We do not yet evaluate broadly on GUI agents~\citep{xie2024osworld,bonatti2024windows}, browser tasks~\citep{zhou2024webarena,koh2024visualwebarena}, mobile agents~\citep{deng2024mobile,rawles2025androidworld}, embodied robotics~\citep{shridhar2020alfred,yang2025embodiedbench}, or persistent multi-session assistants~\citep{wu2024longmemeval,maharana2024evaluating}. Extending the same contextual reasoning framework to these settings is an important future direction.

Current public evaluation still under-measures many aspects of multimodal agency that motivate this work, such as recursive evidence gathering, memory management, verification quality, and closed-loop decision making over long visual streams. Meanwhile, the broader benchmark emphasis of the field has shifted strongly toward coding, browser, and general tool-use agents. We hope future benchmarks will better reflect visually grounded long-horizon reasoning as a distinct capability axis.

\paragraph{Future Directions.}
We see several promising directions for follow-up work. First, the idea of adapting pretrained attention into new efficient forms could be explored beyond GQA-to-MLA conversion, including hybrid, hierarchical, linear, or hardware-specialized attention designs. Second, MCR could be tested in broader multimodal agent settings such as GUI interaction, mobile agents, and embodied systems. Third, contextual reasoning should be studied together with stronger predictive world models, enabling agents that not only gather and organize evidence, but also simulate and compare future outcomes. Finally, we believe there is continuing value in \emph{resource-efficient adaptation}: a large part of the community does not train frontier-scale models from scratch, but instead works on making existing open weights more useful for new tasks, domains, and deployment constraints. We hope this work contributes to that broader effort.

\clearpage
\bibliographystyle{plainnat}
\bibliography{ref}

@String(CVPR= {IEEE Conf. Comput. Vis. Pattern Recog.})

@String(ICCV= {Int. Conf. Comput. Vis.})

@String(ECCV= {Eur. Conf. Comput. Vis.})

@String(CVPR  = {CVPR})

@String(ICCV  = {ICCV})

@String(ECCV  = {ECCV})

@article{vjepa21,
  title={V-jepa 2.1: Unlocking dense features in video self-supervised learning},
  author={Mur-Labadia, Lorenzo and Muckley, Matthew and Bar, Amir and Assran, Mido and Sinha, Koustuv and Rabbat, Mike and LeCun, Yann and Ballas, Nicolas and Bardes, Adrien},
  journal={arXiv preprint arXiv:2603.14482},
  year={2026}
}

@article{vjepa2,
  title={V-jepa 2: Self-supervised video models enable understanding, prediction and planning},
  author={Assran, Mido and Bardes, Adrien and Fan, David and Garrido, Quentin and Howes, Russell and Muckley, Matthew and Rizvi, Ammar and Roberts, Claire and Sinha, Koustuv and Zholus, Artem and others},
  journal={arXiv preprint arXiv:2506.09985},
  year={2025}
}

@article{vjepa,
  title={Revisiting Feature Prediction for Learning Visual Representations from Video},
  author={Bardes, Adrien and Garrido, Quentin and Ponce, Jean and Chen, Xinlei and Rabbat, Michael and LeCun, Yann and Assran, Mido and Ballas, Nicolas},
  journal={arXiv preprint arXiv:2404.08471},
  year={2024}
}

@inproceedings{ijepa,
  title={Self-supervised learning from images with a joint-embedding predictive architecture},
  author={Assran, Mahmoud and Duval, Quentin and Misra, Ishan and Bojanowski, Piotr and Vincent, Pascal and Rabbat, Michael and LeCun, Yann and Ballas, Nicolas},
  booktitle={Proceedings of the IEEE/CVF conference on computer vision and pattern recognition},
  pages={15619--15629},
  year={2023}
}

@article{ha2018world,
  title={World models},
  author={Ha, David and Schmidhuber, J{\"u}rgen},
  journal={arXiv preprint arXiv:1803.10122},
  volume={2},
  number={3},
  pages={440},
  year={2018}
}

@article{keyevl15,
  title={Kwai keye-vl 1.5 technical report},
  author={Yang, Biao and Wen, Bin and Ding, Boyang and Liu, Changyi and Chu, Chenglong and Song, Chengru and Rao, Chongling and Yi, Chuan and Li, Da and Zang, Dunju and others},
  journal={arXiv preprint arXiv:2509.01563},
  year={2025}
}

@misc{sonnet45,
  author       = {{Anthropic}},
  title        = {System Card},
  year         = {2024},
  howpublished = {\url{https://www-cdn.anthropic.com/963373e433e489a87a10c823c52a0a013e9172dd.pdf}},
  note         = {Anthropic PDF document. Accessed: 2026-05-28}
}

@misc{gemini3pro,
  author       = {{Google DeepMind}},
  title        = {Gemini 3 Pro Model Card},
  year         = {2025},
  howpublished = {\url{https://storage.googleapis.com/deepmind-media/Model-Cards/Gemini-3-Pro-Model-Card.pdf}},
  note         = {Model card. Accessed: 2026-05-28}
}

@article{gpt5,
  title={Openai gpt-5 system card},
  author={Singh, Aaditya and Fry, Adam and Perelman, Adam and Tart, Adam and Ganesh, Adi and El-Kishky, Ahmed and McLaughlin, Aidan and Low, Aiden and Ostrow, AJ and Ananthram, Akhila and others},
  journal={arXiv preprint arXiv:2601.03267},
  year={2025}
}

@article{comanici2025gemini,
  title={Gemini 2.5: Pushing the frontier with advanced reasoning, multimodality, long context, and next generation agentic capabilities},
  author={Comanici, Gheorghe and Bieber, Eric and Schaekermann, Mike and Pasupat, Ice and Sachdeva, Noveen and Dhillon, Inderjit and Blistein, Marcel and Ram, Ori and Zhang, Dan and Rosen, Evan and others},
  journal={arXiv preprint arXiv:2507.06261},
  year={2025}
}

@inproceedings{vstar,
  title={V*: Guided visual search as a core mechanism in multimodal llms},
  author={Wu, Penghao and Xie, Saining},
  booktitle={Proceedings of the IEEE/CVF Conference on Computer Vision and Pattern Recognition},
  pages={13084--13094},
  year={2024}
}

@article{tot,
  title={Tree of thoughts: Deliberate problem solving with large language models},
  author={Yao, Shunyu and Yu, Dian and Zhao, Jeffrey and Shafran, Izhak and Griffiths, Tom and Cao, Yuan and Narasimhan, Karthik},
  journal={Advances in neural information processing systems},
  volume={36},
  pages={11809--11822},
  year={2023}
}

@article{cot,
  title={Chain-of-thought prompting elicits reasoning in large language models},
  author={Wei, Jason and Wang, Xuezhi and Schuurmans, Dale and Bosma, Maarten and Xia, Fei and Chi, Ed and Le, Quoc V and Zhou, Denny and others},
  journal={Advances in neural information processing systems},
  volume={35},
  pages={24824--24837},
  year={2022}
}

@inproceedings{react,
  title={React: Synergizing reasoning and acting in language models},
  author={Yao, Shunyu and Zhao, Jeffrey and Yu, Dian and Du, Nan and Shafran, Izhak and Narasimhan, Karthik R and Cao, Yuan},
  booktitle={The eleventh international conference on learning representations},
  year={2022}
}

@inproceedings{moviechat,
  title={Moviechat: From dense token to sparse memory for long video understanding},
  author={Song, Enxin and Chai, Wenhao and Wang, Guanhong and Zhang, Yucheng and Zhou, Haoyang and Wu, Feiyang and Chi, Haozhe and Guo, Xun and Ye, Tian and Zhang, Yanting and others},
  booktitle={Proceedings of the IEEE/CVF Conference on Computer Vision and Pattern Recognition},
  pages={18221--18232},
  year={2024}
}

@inproceedings{videoagent2,
  title={Videoagent: A memory-augmented multimodal agent for video understanding},
  author={Fan, Yue and Ma, Xiaojian and Wu, Rujie and Du, Yuntao and Li, Jiaqi and Gao, Zhi and Li, Qing},
  booktitle={European Conference on Computer Vision},
  pages={75--92},
  year={2024},
  organization={Springer}
}

@inproceedings{videoagent1,
  title={Videoagent: Long-form video understanding with large language model as agent},
  author={Wang, Xiaohan and Zhang, Yuhui and Zohar, Orr and Yeung-Levy, Serena},
  booktitle={European Conference on Computer Vision},
  pages={58--76},
  year={2024},
  organization={Springer}
}

@article{mobileagent,
  title={Mobile-agent: Autonomous multi-modal mobile device agent with visual perception},
  author={Wang, Junyang and Xu, Haiyang and Ye, Jiabo and Yan, Ming and Shen, Weizhou and Zhang, Ji and Huang, Fei and Sang, Jitao},
  journal={arXiv preprint arXiv:2401.16158},
  year={2024}
}

@inproceedings{appagent,
  title={Appagent: Multimodal agents as smartphone users},
  author={Zhang, Chi and Yang, Zhao and Liu, Jiaxuan and Li, Yanda and Han, Yucheng and Chen, Xin and Huang, Zebiao and Fu, Bin and Yu, Gang},
  booktitle={Proceedings of the 2025 CHI Conference on Human Factors in Computing Systems},
  pages={1--20},
  year={2025}
}

@inproceedings{cogagent,
  title={Cogagent: A visual language model for gui agents},
  author={Hong, Wenyi and Wang, Weihan and Lv, Qingsong and Xu, Jiazheng and Yu, Wenmeng and Ji, Junhui and Wang, Yan and Wang, Zihan and Dong, Yuxiao and Ding, Ming and others},
  booktitle={Proceedings of the IEEE/CVF Conference on Computer Vision and Pattern Recognition},
  pages={14281--14290},
  year={2024}
}

@inproceedings{visprog,
  title={Visual programming: Compositional visual reasoning without training},
  author={Gupta, Tanmay and Kembhavi, Aniruddha},
  booktitle={Proceedings of the IEEE/CVF conference on computer vision and pattern recognition},
  pages={14953--14962},
  year={2023}
}

@article{zhang2024llava,
  title={Llava-video: Video instruction tuning with synthetic data},
  author={Zhang, Yuanhan and Wu, Jinming and Li, Wei and Li, Bo and Ma, Zejun and Liu, Ziwei and Li, Chunyuan},
  journal={arXiv preprint arXiv:2410.02713},
  year={2024}
}

@article{Maaz2024VideoGPT+,
    title={VideoGPT+: Integrating Image and Video Encoders for Enhanced Video Understanding},
    author={Maaz, Muhammad and Rasheed, Hanoona and Khan, Salman and Khan, Fahad Shahbaz},
    journal={arxiv},
    year={2024},
    url={https://arxiv.org/abs/2406.09418}
}

@article{wiedmann2025finevision,
  title={Finevision: Open data is all you need},
  author={Wiedmann, Luis and Zohar, Orr and Mahla, Amir and Wang, Xiaohan and Li, Rui and Frere, Thibaud and von Werra, Leandro and Gosthipaty, Aritra Roy and Marafioti, Andr{\'e}s},
  journal={arXiv preprint arXiv:2510.17269},
  year={2025}
}

@article{wang2025internvideo2,
  title={Internvideo2. 5: Empowering video mllms with long and rich context modeling},
  author={Wang, Yi and Li, Xinhao and Yan, Ziang and He, Yinan and Yu, Jiashuo and Zeng, Xiangyu and Wang, Chenting and Ma, Changlian and Huang, Haian and Gao, Jianfei and others},
  journal={arXiv preprint arXiv:2501.12386},
  year={2025}
}

@article{clark2026molmo2,
  title={Molmo2: Open Weights and Data for Vision-Language Models with Video Understanding and Grounding},
  author={Clark, Christopher and Zhang, Jieyu and Ma, Zixian and Park, Jae Sung and Salehi, Mohammadreza and Tripathi, Rohun and Lee, Sangho and Ren, Zhongzheng and Kim, Chris Dongjoo and Yang, Yinuo and others},
  journal={arXiv preprint arXiv:2601.10611},
  year={2026}
}

@article{yu2025minicpm,
  title={Minicpm-v 4.5: Cooking efficient mllms via architecture, data, and training recipe},
  author={Yu, Tianyu and Wang, Zefan and Wang, Chongyi and Huang, Fuwei and Ma, Wenshuo and He, Zhihui and Cai, Tianchi and Chen, Weize and Huang, Yuxiang and Zhao, Yuanqian and others},
  journal={arXiv preprint arXiv:2509.18154},
  year={2025}
}

@misc{qwen35blog,
    title = {Qwen3.5: Accelerating Productivity with Native Multimodal Agents},
    url = {https://qwen.ai/blog?id=qwen3.5},
    author = {Qwen Team},
    month = {February},
    year = {2026}
}

@misc{deepseekai2026deepseekv4,
      title={DeepSeek-V4: Towards Highly Efficient Million-Token Context Intelligence},
      author={DeepSeek-AI},
      year={2026},
}

@article{zeng2026glm,
  title={Glm-5: from vibe coding to agentic engineering},
  author={Zeng, Aohan and Lv, Xin and Hou, Zhenyu and Du, Zhengxiao and Zheng, Qinkai and Chen, Bin and Yin, Da and Ge, Chendi and Huang, Chenghua and Xie, Chengxing and others},
  journal={arXiv preprint arXiv:2602.15763},
  year={2026}
}

@article{huang2026step,
  title={Step 3.5 Flash: Open Frontier-Level Intelligence with 11B Active Parameters},
  author={Huang, Ailin and Li, Ang and Kong, Aobo and Wang, Bin and Jiao, Binxing and Dong, Bo and Wang, Bojun and Chen, Boyu and Li, Brian and Ma, Buyun and others},
  journal={arXiv preprint arXiv:2602.10604},
  year={2026}
}

@misc{minimax2026m3,
  author       = {{MiniMax Research}},
  title        = {MiniMax M3: Frontier Coding Capabilities, 1M Context, and Native Multimodality in One Model},
  year         = {2026},
  month        = jun,
  day          = {1},
  howpublished = {\url{https://www.minimaxi.com/blog/minimax-m3}},
  note         = {MiniMax blog, accessed 2026-06-05}
}

@misc{meituanlongcatteam2026longcatnextlexicalizingmodalitiesdiscrete,
      title={LongCat-Next: Lexicalizing Modalities as Discrete Tokens}, 
      author={Meituan LongCat Team},
      year={2026},
      eprint={2603.27538},
      archivePrefix={arXiv},
      primaryClass={cs.CV},
      url={https://arxiv.org/abs/2603.27538}, 
}

@article{liu2024deepseek,
  title={Deepseek-v2: A strong, economical, and efficient mixture-of-experts language model},
  author={Liu, Aixin and Feng, Bei and Wang, Bin and Wang, Bingxuan and Liu, Bo and Zhao, Chenggang and Dengr, Chengqi and Ruan, Chong and Dai, Damai and Guo, Daya and others},
  journal={arXiv preprint arXiv:2405.04434},
  year={2024}
}

@article{guo2025deepseek,
  title={Deepseek-r1: Incentivizing reasoning capability in llms via reinforcement learning},
  author={Guo, Daya and Yang, Dejian and Zhang, Haowei and Song, Junxiao and Wang, Peiyi and Zhu, Qihao and Xu, Runxin and Zhang, Ruoyu and Ma, Shirong and Bi, Xiao and others},
  journal={arXiv preprint arXiv:2501.12948},
  year={2025}
}

@article{team2025kimi,
  title={Kimi linear: An expressive, efficient attention architecture},
  author={Team, Kimi and Zhang, Yu and Lin, Zongyu and Yao, Xingcheng and Hu, Jiaxi and Meng, Fanqing and Liu, Chengyin and Men, Xin and Yang, Songlin and Li, Zhiyuan and others},
  journal={arXiv preprint arXiv:2510.26692},
  year={2025}
}

@inproceedings{zhang2026timelens,
  title={Timelens: Rethinking video temporal grounding with multimodal llms},
  author={Zhang, Jun and Wang, Teng and Ge, Yuying and Ge, Yixiao and Li, Xinhao and Wang, Limin},
  booktitle={Proceedings of the IEEE/CVF Conference on Computer Vision and Pattern Recognition},
  pages={10419--10429},
  year={2026}
}

@article{team2026kimi,
  title={Kimi K2. 5: Visual Agentic Intelligence},
  author={Team, Kimi and Bai, Tongtong and Bai, Yifan and Bao, Yiping and Cai, SH and Cao, Yuan and Charles, Y and Che, HS and Chen, Cheng and Chen, Guanduo and others},
  journal={arXiv preprint arXiv:2602.02276},
  year={2026}
}

@article{chen2026eagle,
  title={Eagle 2.5: Boosting long-context post-training for frontier vision-language models},
  author={Chen, Guo and Li, Zhiqi and Wang, Shihao and Jiang, Jindong and Liu, Yicheng and Lu, Lidong and Huang, De-An and Byeon, Wonmin and Le, Matthieu and Ehrlich, Max and others},
  journal={Advances in Neural Information Processing Systems},
  volume={38},
  pages={91077--91100},
  year={2026}
}

@inproceedings{vipergpt,
  title={Vipergpt: Visual inference via python execution for reasoning},
  author={Sur{\'\i}s, D{\'\i}dac and Menon, Sachit and Vondrick, Carl},
  booktitle={Proceedings of the IEEE/CVF international conference on computer vision},
  pages={11888--11898},
  year={2023}
}

@misc{qwen3vl,
      title={Qwen3-VL Technical Report}, 
      author={Shuai Bai and Yuxuan Cai and Ruizhe Chen and Keqin Chen and Xionghui Chen and Zesen Cheng and Lianghao Deng and Wei Ding and Chang Gao and Chunjiang Ge and Wenbin Ge and Zhifang Guo and Qidong Huang and Jie Huang and Fei Huang and Binyuan Hui and Shutong Jiang and Zhaohai Li and Mingsheng Li and Mei Li and Kaixin Li and Zicheng Lin and Junyang Lin and Xuejing Liu and Jiawei Liu and Chenglong Liu and Yang Liu and Dayiheng Liu and Shixuan Liu and Dunjie Lu and Ruilin Luo and Chenxu Lv and Rui Men and Lingchen Meng and Xuancheng Ren and Xingzhang Ren and Sibo Song and Yuchong Sun and Jun Tang and Jianhong Tu and Jianqiang Wan and Peng Wang and Pengfei Wang and Qiuyue Wang and Yuxuan Wang and Tianbao Xie and Yiheng Xu and Haiyang Xu and Jin Xu and Zhibo Yang and Mingkun Yang and Jianxin Yang and An Yang and Bowen Yu and Fei Zhang and Hang Zhang and Xi Zhang and Bo Zheng and Humen Zhong and Jingren Zhou and Fan Zhou and Jing Zhou and Yuanzhi Zhu and Ke Zhu},
      year={2025},
      eprint={2511.21631},
      archivePrefix={arXiv},
      primaryClass={cs.CV},
      url={https://arxiv.org/abs/2511.21631}, 
}

@article{qwen25vl,
  title={Qwen2. 5-vl technical report},
  author={Bai, Shuai and Chen, Keqin and Liu, Xuejing and Wang, Jialin and Ge, Wenbin and Song, Sibo and Dang, Kai and Wang, Peng and Wang, Shijie and Tang, Jun and others},
  journal={arXiv preprint arXiv:2502.13923},
  year={2025}
}

@inproceedings{clip,
  title={Learning transferable visual models from natural language supervision},
  author={Radford, Alec and Kim, Jong Wook and Hallacy, Chris and Ramesh, Aditya and Goh, Gabriel and Agarwal, Sandhini and Sastry, Girish and Askell, Amanda and Mishkin, Pamela and Clark, Jack and others},
  booktitle={International conference on machine learning},
  pages={8748--8763},
  year={2021},
  organization={PmLR}
}

@misc{mimov25,
  title={MiMo-V2.5},
  year={2026},
  howpublished={\url{https://huggingface.co/collections/XiaomiMiMo/mimo-v25}},
}

@article{flamingo,
  title={Flamingo: a visual language model for few-shot learning},
  author={Alayrac, Jean-Baptiste and Donahue, Jeff and Luc, Pauline and Miech, Antoine and Barr, Iain and Hasson, Yana and Lenc, Karel and Mensch, Arthur and Millican, Katherine and Reynolds, Malcolm and others},
  journal={Advances in neural information processing systems},
  volume={35},
  pages={23716--23736},
  year={2022}
}

@article{zhang2025dsi,
  title={Dsi-bench: A benchmark for dynamic spatial intelligence},
  author={Zhang, Ziang and Wang, Zehan and Zhang, Guanghao and Dai, Weilong and Xia, Yan and Yan, Ziang and Hong, Minjie and Zhao, Zhou},
  journal={arXiv preprint arXiv:2510.18873},
  year={2025}
}

@article{yang2025mmsi,
  title={Mmsi-bench: A benchmark for multi-image spatial intelligence},
  author={Yang, Sihan and Xu, Runsen and Xie, Yiman and Yang, Sizhe and Li, Mo and Lin, Jingli and Zhu, Chenming and Chen, Xiaochen and Duan, Haodong and Yue, Xiangyu and others},
  journal={arXiv preprint arXiv:2505.23764},
  year={2025}
}

@inproceedings{zheng2025deepresearcher,
  title={Deepresearcher: Scaling deep research via reinforcement learning in real-world environments},
  author={Zheng, Yuxiang and Fu, Dayuan and Hu, Xiangkun and Cai, Xiaojie and Ye, Lyumanshan and Lu, Pengrui and Liu, Pengfei},
  booktitle={Proceedings of the 2025 Conference on Empirical Methods in Natural Language Processing},
  pages={414--431},
  year={2025}
}

@article{yang2024swe,
  title={Swe-agent: Agent-computer interfaces enable automated software engineering},
  author={Yang, John and Jimenez, Carlos E and Wettig, Alexander and Lieret, Kilian and Yao, Shunyu and Narasimhan, Karthik and Press, Ofir},
  journal={Advances in Neural Information Processing Systems},
  volume={37},
  pages={50528--50652},
  year={2024}
}

@inproceedings{wang2025openhands,
  title={Openhands: An open platform for ai software developers as generalist agents},
  author={Wang, Xingyao and Li, Boxuan and Song, Yufan and Xu, Frank F and Tang, Xiangru and Zhuge, Mingchen and Pan, Jiayi and Song, Yueqi and Li, Bowen and Singh, Jaskirat and others},
  booktitle={International Conference on Learning Representations},
  volume={2025},
  pages={65882--65919},
  year={2025}
}

@article{yao2022webshop,
  title={Webshop: Towards scalable real-world web interaction with grounded language agents},
  author={Yao, Shunyu and Chen, Howard and Yang, John and Narasimhan, Karthik},
  journal={Advances in Neural Information Processing Systems},
  volume={35},
  pages={20744--20757},
  year={2022}
}

@inproceedings{zhou2024webarena,
  title={Webarena: A realistic web environment for building autonomous agents},
  author={Zhou, Shuyan and Xu, Frank F and Zhu, Hao and Zhou, Xuhui and Lo, Robert and Sridhar, Abishek and Cheng, Xianyi and Ou, Tianyue and Bisk, Yonatan and Fried, Daniel and others},
  booktitle={International Conference on Learning Representations},
  volume={2024},
  pages={15585--15606},
  year={2024}
}

@inproceedings{koh2024visualwebarena,
  title={Visualwebarena: Evaluating multimodal agents on realistic visual web tasks},
  author={Koh, Jing Yu and Lo, Robert and Jang, Lawrence and Duvvur, Vikram and Lim, Ming and Huang, Po-Yu and Neubig, Graham and Zhou, Shuyan and Salakhutdinov, Russ and Fried, Daniel},
  booktitle={Proceedings of the 62nd Annual Meeting of the Association for Computational Linguistics (Volume 1: Long Papers)},
  pages={881--905},
  year={2024}
}

@inproceedings{qin2024toolllm,
  title={Toolllm: Facilitating large language models to master 16000+ real-world apis},
  author={Qin, Yujia and Liang, Shihao and Ye, Yining and Zhu, Kunlun and Yan, Lan and Lu, Yaxi and Lin, Yankai and Cong, Xin and Tang, Xiangru and Qian, Bill and others},
  booktitle={International Conference on Learning Representations},
  volume={2024},
  pages={9695--9717},
  year={2024}
}

@article{jin2025search,
  title={Search-r1: Training llms to reason and leverage search engines with reinforcement learning},
  author={Jin, Bowen and Zeng, Hansi and Yue, Zhenrui and Yoon, Jinsung and Arik, Sercan and Wang, Dong and Zamani, Hamed and Han, Jiawei},
  journal={arXiv preprint arXiv:2503.09516},
  year={2025}
}

@misc{penedo2026finetranslations,
      title={FineTranslations}, 
      author={Guilherme Penedo and Hynek Kydl{\'\i}{\v{c}}ek and Amir Hossein Kargaran and Leandro von Werra},
      year={2026},
      publisher = {Hugging Face},
      journal = {Hugging Face repository},
      howpublished = {\url{https://huggingface.co/datasets/HuggingFaceFW/finetranslations}}
}

@article{lin2025mmsi,
  title={MMSI-Video-Bench: A Holistic Benchmark for Video-Based Spatial Intelligence},
  author={Lin, Jingli and Xu, Runsen and Zhu, Shaohao and Yang, Sihan and Cao, Peizhou and Ran, Yunlong and Hu, Miao and Zhu, Chenming and Xie, Yiman and Long, Yilin and others},
  journal={arXiv preprint arXiv:2512.10863},
  year={2025}
}

@inproceedings{yang2025thinking,
  title={Thinking in space: How multimodal large language models see, remember, and recall spaces},
  author={Yang, Jihan and Yang, Shusheng and Gupta, Anjali W and Han, Rilyn and Fei-Fei, Li and Xie, Saining},
  booktitle={Proceedings of the Computer Vision and Pattern Recognition Conference},
  pages={10632--10643},
  year={2025}
}

@article{internvl35,
  title={Internvl3.5: Advancing open-source multimodal models in versatility, reasoning, and efficiency},
  author={Wang, Weiyun and Gao, Zhangwei and Gu, Lixin and Pu, Hengjun and Cui, Long and Wei, Xingguang and Liu, Zhaoyang and Jing, Linglin and Ye, Shenglong and Shao, Jie and others},
  journal={arXiv preprint arXiv:2508.18265},
  year={2025}
}

@inproceedings{pmlr-v202-li23q,
  title={Blip-2: Bootstrapping language-image pre-training with frozen image encoders and large language models},
  author={Li, Junnan and Li, Dongxu and Savarese, Silvio and Hoi, Steven},
  booktitle={ICML},
  pages={19730--19742},
  year={2023},
  organization={PMLR}
}

@article{su2024roformer,
  title={Roformer: Enhanced transformer with rotary position embedding},
  author={Su, Jianlin and Ahmed, Murtadha and Lu, Yu and Pan, Shengfeng and Bo, Wen and Liu, Yunfeng},
  journal={Neurocomputing},
  volume={568},
  pages={127063},
  year={2024},
  publisher={Elsevier}
}

@article{zhang2019root,
  title={Root mean square layer normalization},
  author={Zhang, Biao and Sennrich, Rico},
  journal={Advances in neural information processing systems},
  volume={32},
  year={2019}
}

@inproceedings{henry2020query,
  title={Query-key normalization for transformers},
  author={Henry, Alex and Dachapally, Prudhvi Raj and Pawar, Shubham Shantaram and Chen, Yuxuan},
  booktitle={Findings of the Association for Computational Linguistics: EMNLP 2020},
  pages={4246--4253},
  year={2020}
}

@inproceedings{ainslie2023gqa,
  title={Gqa: Training generalized multi-query transformer models from multi-head checkpoints},
  author={Ainslie, Joshua and Lee-Thorp, James and De Jong, Michiel and Zemlyanskiy, Yury and Lebr{\'o}n, Federico and Sanghai, Sumit},
  booktitle={Proceedings of the 2023 Conference on Empirical Methods in Natural Language Processing},
  pages={4895--4901},
  year={2023}
}

@inproceedings{sultani2018real,
  title={Real-world anomaly detection in surveillance videos},
  author={Sultani, Waqas and Chen, Chen and Shah, Mubarak},
  booktitle={Proceedings of the IEEE conference on computer vision and pattern recognition},
  pages={6479--6488},
  year={2018}
}

@article{hansen2022temporal,
  title={Temporal difference learning for model predictive control},
  author={Hansen, Nicklas and Wang, Xiaolong and Su, Hao},
  journal={arXiv preprint arXiv:2203.04955},
  year={2022}
}

@inproceedings{hansen2024td,
  title={Td-mpc2: Scalable, robust world models for continuous control},
  author={Hansen, Nick and Su, Hao and Wang, Xiaolong},
  booktitle={International Conference on Learning Representations},
  volume={2024},
  pages={47376--47405},
  year={2024}
}

@article{machado2023temporal,
  title={Temporal abstraction in reinforcement learning with the successor representation},
  author={Machado, Marlos C and Barreto, Andre and Precup, Doina and Bowling, Michael},
  journal={Journal of machine learning research},
  volume={24},
  number={80},
  pages={1--69},
  year={2023}
}

@article{kong2024latent,
  title={Latent plan transformer for trajectory abstraction: Planning as latent space inference},
  author={Kong, Deqian and Xu, Dehong and Zhao, Minglu and Pang, Bo and Xie, Jianwen and Lizarraga, Andrew and Huang, Yuhao and Xie, Sirui and Wu, Ying Nian},
  journal={Advances in Neural Information Processing Systems},
  volume={37},
  pages={123379--123401},
  year={2024}
}

@inproceedings{asai2024self,
  title={Self-rag: Learning to retrieve, generate, and critique through self-reflection},
  author={Asai, Akari and Wu, Zeqiu and Wang, Yizhong and Sil, Avi and Hajishirzi, Hannaneh},
  booktitle={International conference on learning representations},
  volume={2024},
  pages={9112--9141},
  year={2024}
}

@article{ahn2022can,
  title={Do as i can, not as i say: Grounding language in robotic affordances},
  author={Ahn, Michael and Brohan, Anthony and Brown, Noah and Chebotar, Yevgen and Cortes, Omar and David, Byron and Finn, Chelsea and Fu, Chuyuan and Gopalakrishnan, Keerthana and Hausman, Karol and others},
  journal={arXiv preprint arXiv:2204.01691},
  year={2022}
}

@article{brohan2022rt,
  title={Rt-1: Robotics transformer for real-world control at scale},
  author={Brohan, Anthony and Brown, Noah and Carbajal, Justice and Chebotar, Yevgen and Dabis, Joseph and Finn, Chelsea and Gopalakrishnan, Keerthana and Hausman, Karol and Herzog, Alex and Hsu, Jasmine and others},
  journal={arXiv preprint arXiv:2212.06817},
  year={2022}
}

@inproceedings{agarwal2024policy,
  title={On-policy distillation of language models: Learning from self-generated mistakes},
  author={Agarwal, Rishabh and Vieillard, Nino and Zhou, Yongchao and Stanczyk, Piotr and Ramos Garea, Sabela and Geist, Matthieu and Bachem, Olivier},
  booktitle={International Conference on Learning Representations},
  volume={2024},
  pages={21246--21263},
  year={2024}
}

@article{lu2025onpolicydistillation,
  author = {Kevin Lu and Thinking Machines Lab},
  title = {On-Policy Distillation},
  journal = {Thinking Machines Lab: Connectionism},
  year = {2025},
  note = {https://thinkingmachines.ai/blog/on-policy-distillation},
  doi = {10.64434/tml.20251026},
}

@inproceedings{xu2025speculative,
  title={Speculative knowledge distillation: Bridging the teacher-student gap through interleaved sampling},
  author={Xu, Wenda and Han, Rujun and Wang, Zifeng and Le, Long and Madeka, Dhruv and Li, Lei and Wang, William and Agarwal, Rishabh and Lee, Chen-Yu and Pfister, Tomas},
  booktitle={International Conference on Learning Representations},
  volume={2025},
  pages={64616--64646},
  year={2025}
}

@article{schick2023toolformer,
  title={Toolformer: Language models can teach themselves to use tools},
  author={Schick, Timo and Dwivedi-Yu, Jane and Dess{\`\i}, Roberto and Raileanu, Roberta and Lomeli, Maria and Hambro, Eric and Zettlemoyer, Luke and Cancedda, Nicola and Scialom, Thomas},
  journal={Advances in neural information processing systems},
  volume={36},
  pages={68539--68551},
  year={2023}
}

@inproceedings{bai2025longbench,
  title={Longbench v2: Towards deeper understanding and reasoning on realistic long-context multitasks},
  author={Bai, Yushi and Tu, Shangqing and Zhang, Jiajie and Peng, Hao and Wang, Xiaozhi and Lv, Xin and Cao, Shulin and Xu, Jiazheng and Hou, Lei and Dong, Yuxiao and others},
  booktitle={Proceedings of the 63rd Annual Meeting of the Association for Computational Linguistics (Volume 1: Long Papers)},
  pages={3639--3664},
  year={2025}
}

@inproceedings{zhang2024bench,
  title={{$\infty$ Bench}: Extending long context evaluation beyond 100K tokens},
  author={Zhang, Xinrong and Chen, Yingfa and Hu, Shengding and Xu, Zihang and Chen, Junhao and Hao, Moo and Han, Xu and Thai, Zhen and Wang, Shuo and Liu, Zhiyuan and others},
  booktitle={Proceedings of the 62nd Annual Meeting of the Association for Computational Linguistics (Volume 1: Long Papers)},
  pages={15262--15277},
  year={2024}
}

@inproceedings{park2023generative,
  title={Generative agents: Interactive simulacra of human behavior},
  author={Park, Joon Sung and O'Brien, Joseph and Cai, Carrie Jun and Morris, Meredith Ringel and Liang, Percy and Bernstein, Michael S},
  booktitle={Proceedings of the 36th annual acm symposium on user interface software and technology},
  pages={1--22},
  year={2023}
}

@article{shinn2023reflexion,
  title={Reflexion: Language agents with verbal reinforcement learning},
  author={Shinn, Noah and Cassano, Federico and Gopinath, Ashwin and Narasimhan, Karthik and Yao, Shunyu},
  journal={Advances in neural information processing systems},
  volume={36},
  pages={8634--8652},
  year={2023}
}

@article{yang2025qwen3,
  title={Qwen3 technical report},
  author={Yang, An and Li, Anfeng and Yang, Baosong and Zhang, Beichen and Hui, Binyuan and Zheng, Bo and Yu, Bowen and Gao, Chang and Huang, Chengen and Lv, Chenxu and others},
  journal={arXiv preprint arXiv:2505.09388},
  year={2025}
}

@article{hafner2019dream,
  title={Dream to control: Learning behaviors by latent imagination},
  author={Hafner, Danijar and Lillicrap, Timothy and Ba, Jimmy and Norouzi, Mohammad},
  journal={arXiv preprint arXiv:1912.01603},
  year={2019}
}

@article{hafner2023mastering,
  title={Mastering diverse domains through world models},
  author={Hafner, Danijar and Pasukonis, Jurgis and Ba, Jimmy and Lillicrap, Timothy},
  journal={arXiv preprint arXiv:2301.04104},
  year={2023}
}

@article{xie2024osworld,
  title={Osworld: Benchmarking multimodal agents for open-ended tasks in real computer environments},
  author={Xie, Tianbao and Zhang, Danyang and Chen, Jixuan and Li, Xiaochuan and Zhao, Siheng and Cao, Ruisheng and Hua, Toh J and Cheng, Zhoujun and Shin, Dongchan and Lei, Fangyu and others},
  journal={Advances in Neural Information Processing Systems},
  volume={37},
  pages={52040--52094},
  year={2024}
}

@article{bonatti2024windows,
  title={Windows agent arena: Evaluating multi-modal os agents at scale},
  author={Bonatti, Rogerio and Zhao, Dan and Bonacci, Francesco and Dupont, Dillon and Abdali, Sara and Li, Yinheng and Lu, Yadong and Wagle, Justin and Koishida, Kazuhito and Bucker, Arthur and others},
  journal={arXiv preprint arXiv:2409.08264},
  year={2024}
}

@inproceedings{deng2024mobile,
  title={Mobile-bench: An evaluation benchmark for llm-based mobile agents},
  author={Deng, Shihan and Xu, Weikai and Sun, Hongda and Liu, Wei and Tan, Tao and Liujianfeng, Liujianfeng and Li, Ang and Luan, Jian and Wang, Bin and Yan, Rui and others},
  booktitle={Proceedings of the 62nd Annual Meeting of the Association for Computational Linguistics (Volume 1: Long Papers)},
  pages={8813--8831},
  year={2024}
}

@article{wu2024longmemeval,
  title={Longmemeval: Benchmarking chat assistants on long-term interactive memory},
  author={Wu, Di and Wang, Hongwei and Yu, Wenhao and Zhang, Yuwei and Chang, Kai-Wei and Yu, Dong},
  journal={arXiv preprint arXiv:2410.10813},
  year={2024}
}

@inproceedings{maharana2024evaluating,
  title={Evaluating very long-term conversational memory of llm agents},
  author={Maharana, Adyasha and Lee, Dong-Ho and Tulyakov, Sergey and Bansal, Mohit and Barbieri, Francesco and Fang, Yuwei},
  booktitle={Proceedings of the 62nd Annual Meeting of the Association for Computational Linguistics (Volume 1: Long Papers)},
  pages={13851--13870},
  year={2024}
}

@inproceedings{rawles2025androidworld,
  title={Androidworld: A dynamic benchmarking environment for autonomous agents},
  author={Rawles, Chris and Clinckemaillie, Sarah and Chang, Yifan and Waltz, Jonathan and Lau, Gabrielle and Fair, Marybeth and Li, Alice and Bishop, William and Li, Wei and Campbell-Ajala, Folawiyo and others},
  booktitle={International Conference on Learning Representations},
  volume={2025},
  pages={406--441},
  year={2025}
}

@inproceedings{shridhar2020alfred,
  title={Alfred: A benchmark for interpreting grounded instructions for everyday tasks},
  author={Shridhar, Mohit and Thomason, Jesse and Gordon, Daniel and Bisk, Yonatan and Han, Winson and Mottaghi, Roozbeh and Zettlemoyer, Luke and Fox, Dieter},
  booktitle={Proceedings of the IEEE/CVF conference on computer vision and pattern recognition},
  pages={10740--10749},
  year={2020}
}

@article{yang2025embodiedbench,
  title={Embodiedbench: Comprehensive benchmarking multi-modal large language models for vision-driven embodied agents},
  author={Yang, Rui and Chen, Hanyang and Zhang, Junyu and Zhao, Mark and Qian, Cheng and Wang, Kangrui and Wang, Qineng and Koripella, Teja Venkat and Movahedi, Marziyeh and Li, Manling and others},
  journal={arXiv preprint arXiv:2502.09560},
  year={2025}
}

@inproceedings{yang2025cambrian,
  title={Cambrian-s: Towards spatial supersensing in video},
  author={Yang, Shusheng and Yang, Jihan and Huang, Pinzhi and Brown II, Ellis L and Yang, Zihao and Yu, Yue and Tong, Shengbang and Zheng, Zihan and Xu, Yifan and Wang, Muhan and others},
  booktitle={The Fourteenth International Conference on Learning Representations},
  year={2025}
}

@inproceedings{caba2015activitynet,
  title={Activitynet: A large-scale video benchmark for human activity understanding},
  author={Caba Heilbron, Fabian and Escorcia, Victor and Ghanem, Bernard and Carlos Niebles, Juan},
  booktitle={CVPR},
  pages={961--970},
  year={2015}
}

@article{qvhighlight,
  title={Detecting moments and highlights in videos via natural language queries},
  author={Lei, Jie and Berg, Tamara L and Bansal, Mohit},
  journal={NeurIPS},
  volume={34},
  pages={11846--11858},
  year={2021}
}

@inproceedings{mvbench,
  title={Mvbench: A comprehensive multi-modal video understanding benchmark},
  author={Li, Kunchang and Wang, Yali and He, Yinan and Li, Yizhuo and Wang, Yi and Liu, Yi and Wang, Zun and Xu, Jilan and Chen, Guo and Luo, Ping and others},
  booktitle={CVPR},
  pages={22195--22206},
  year={2024}
}

@inproceedings{shangguan2025tomato,
  title={Tomato: Assessing visual temporal reasoning capabilities in multimodal foundation models},
  author={Shangguan, Ziyao and Li, Chuhan and Ding, Yuxuan and Zheng, Yanan and Zhao, Yilun and Fitzgerald, Tesca and Cohan, Arman},
  booktitle={International Conference on Learning Representations},
  volume={2025},
  pages={7593--7734},
  year={2025}
}

@inproceedings{hong2025motionbench,
  title={Motionbench: Benchmarking and improving fine-grained video motion understanding for vision language models},
  author={Hong, Wenyi and Cheng, Yean and Yang, Zhuoyi and Wang, Weihan and Wang, Lefan and Gu, Xiaotao and Huang, Shiyu and Dong, Yuxiao and Tang, Jie},
  booktitle={Proceedings of the Computer Vision and Pattern Recognition Conference},
  pages={8450--8460},
  year={2025}
}

@inproceedings{liu2024tempcompass,
  title={Tempcompass: Do video llms really understand videos?},
  author={Liu, Yuanxin and Li, Shicheng and Liu, Yi and Wang, Yuxiang and Ren, Shuhuai and Li, Lei and Chen, Sishuo and Sun, Xu and Hou, Lu},
  booktitle={Findings of the Association for Computational Linguistics: ACL 2024},
  pages={8731--8772},
  year={2024}
}

@inproceedings{llava,
  title={Visual instruction tuning},
  author={Liu, Haotian and Li, Chunyuan and Wu, Qingyang and Lee, Yong Jae},
  booktitle={NeurIPS},
  volume={36},
  year={2024}
}

@inproceedings{yu2016refcoco,
  title={Modeling context in referring expressions},
  author={Yu, Licheng and Poirson, Patrick and Yang, Shan and Berg, Alexander C and Berg, Tamara L},
  booktitle={ECCV},
  pages={69--85},
  year={2016},
  organization={Springer}
}

@inproceedings{lei2018tvqa,
  title={TVQA: Localized, Compositional Video Question Answering},
  author={Lei, Jie and Yu, Licheng and Bansal, Mohit and Berg, Tamara L},
  booktitle={EMNLP},
  year={2018}
}

@inproceedings{nextgqa,
  title={Can i trust your answer? visually grounded video question answering},
  author={Xiao, Junbin and Yao, Angela and Li, Yicong and Chua, Tat-Seng},
  booktitle={CVPR},
  year={2024}
}

@inproceedings{hudson2019gqa,
  title={Gqa: A new dataset for real-world visual reasoning and compositional question answering},
  author={Hudson, Drew A and Manning, Christopher D},
  booktitle={Proceedings of the IEEE/CVF conference on computer vision and pattern recognition},
  pages={6700--6709},
  year={2019}
}

@article{hong2025glm,
  title={Glm-4.5 v and glm-4.1 v-thinking: Towards versatile multimodal reasoning with scalable reinforcement learning},
  author={Hong, Wenyi and Yu, Wenmeng and Gu, Xiaotao and Wang, Guo and Gan, Guobing and Tang, Haomiao and Cheng, Jiale and Qi, Ji and Ji, Junhui and Pan, Lihang and others},
  journal={arXiv preprint arXiv:2507.01006},
  year={2025}
}

@article{wang2024lvbench,
  title={Lvbench: An extreme long video understanding benchmark},
  author={Wang, Weihan and He, Zehai and Hong, Wenyi and Cheng, Yean and Zhang, Xiaohan and Qi, Ji and Gu, Xiaotao and Huang, Shiyu and Xu, Bin and Dong, Yuxiao and others},
  journal={arXiv preprint arXiv:2406.08035},
  year={2024}
}

@inproceedings{mathew2021docvqa,
  title={Docvqa: A dataset for vqa on document images},
  author={Mathew, Minesh and Karatzas, Dimosthenis and Jawahar, CV},
  booktitle={Proceedings of the IEEE/CVF winter conference on applications of computer vision},
  pages={2200--2209},
  year={2021}
}

@inproceedings{masry2023unichart,
  title={Unichart: A universal vision-language pretrained model for chart comprehension and reasoning},
  author={Masry, Ahmed and Kavehzadeh, Parsa and Hoque, Enamul and Joty, Shafiq and others},
  booktitle={Proceedings of the 2023 conference on empirical methods in natural language processing},
  pages={14662--14684},
  year={2023}
}

@inproceedings{internvl,
  title={Internvl: Scaling up vision foundation models and aligning for generic visual-linguistic tasks},
  author={Chen, Zhe and Wu, Jiannan and Wang, Wenhai and Su, Weijie and Chen, Guo and Xing, Sen and Zhong, Muyan and Zhang, Qinglong and Zhu, Xizhou and Lu, Lewei and others},
  booktitle={CVPR},
  pages={24185--24198},
  year={2024}
}

@article{li2024llava,
  title={Llava-next-interleave: Tackling multi-image, video, and 3d in large multimodal models},
  author={Li, Feng and Zhang, Renrui and Zhang, Hao and Zhang, Yuanhan and Li, Bo and Li, Wei and Ma, Zejun and Li, Chunyuan},
  journal={arXiv preprint arXiv:2407.07895},
  year={2024}
}

@article{Llava-onevision,
  title={Llava-onevision: Easy visual task transfer},
  author={Li, Bo and Zhang, Yuanhan and Guo, Dong and Zhang, Renrui and Li, Feng and Zhang, Hao and Zhang, Kaichen and Li, Yanwei and Liu, Ziwei and Li, Chunyuan},
  journal={arXiv preprint arXiv:2408.03326},
  year={2024}
}

@article{marti2002iam,
  title={The IAM-database: an English sentence database for offline handwriting recognition},
  author={Marti, U-V and Bunke, Horst},
  journal={International journal on document analysis and recognition},
  volume={5},
  number={1},
  pages={39--46},
  year={2002},
  publisher={Springer}
}

@inproceedings{charades-sta,
  title={Tall: Temporal activity localization via language query},
  author={Gao, Jiyang and Sun, Chen and Yang, Zhenheng and Nevatia, Ram},
  booktitle={ICCV},
  pages={5267--5275},
  year={2017}
}

@article{egoschema,
  title={Egoschema: A diagnostic benchmark for very long-form video language understanding},
  author={Mangalam, Karttikeya and Akshulakov, Raiymbek and Malik, Jitendra},
  journal={Advances in Neural Information Processing Systems},
  volume={36},
  pages={46212--46244},
  year={2023}
}

@inproceedings{xiao2021next,
  title={Next-qa: Next phase of question-answering to explaining temporal actions},
  author={Xiao, Junbin and Shang, Xindi and Yao, Angela and Chua, Tat-Seng},
  booktitle={Proceedings of the IEEE/CVF conference on computer vision and pattern recognition},
  pages={9777--9786},
  year={2021}
}

@article{longvideobench,
  title={Longvideobench: A benchmark for long-context interleaved video-language understanding},
  author={Wu, Haoning and Li, Dongxu and Chen, Bei and Li, Junnan},
  journal={arXiv preprint arXiv:2407.15754},
  year={2024}
}

@article{mlvu,
  title={MLVU: A Comprehensive Benchmark for Multi-Task Long Video Understanding},
  author={Zhou, Junjie and Shu, Yan and Zhao, Bo and Wu, Boya and Xiao, Shitao and Yang, Xi and Xiong, Yongping and Zhang, Bo and Huang, Tiejun and Liu, Zheng},
  journal={arXiv preprint arXiv:2406.04264},
  year={2024}
}

@article{videomme,
  title={Video-MME: The First-Ever Comprehensive Evaluation Benchmark of Multi-modal LLMs in Video Analysis},
  author={Fu, Chaoyou and Dai, Yuhan and Luo, Yondong and Li, Lei and Ren, Shuhuai and Zhang, Renrui and Wang, Zihan and Zhou, Chenyu and Shen, Yunhang and Zhang, Mengdan and others},
  journal={arXiv preprint arXiv:2405.21075},
  year={2024}
}

@article{gemini,
  author       = {Machel Reid and
                  Nikolay Savinov and
                  Denis Teplyashin and
                  Dmitry Lepikhin and
                  Timothy P. Lillicrap and
                  Jean{-}Baptiste Alayrac and
                  Radu Soricut and
                  Angeliki Lazaridou and
                  Orhan Firat and
                  Julian Schrittwieser and
                  Ioannis Antonoglou and
                  Rohan Anil and
                  Sebastian Borgeaud and
                  Andrew M. Dai and
                  Katie Millican and
                  Ethan Dyer and
                  Mia Glaese and
                  Thibault Sottiaux and
                  Benjamin Lee and
                  Fabio Viola and
                  Malcolm Reynolds and
                  Yuanzhong Xu and
                  James Molloy and
                  Jilin Chen and
                  Michael Isard and
                  Paul Barham and
                  Tom Hennigan and
                  Ross McIlroy and
                  Melvin Johnson and
                  Johan Schalkwyk and
                  Eli Collins and
                  Eliza Rutherford and
                  Erica Moreira and
                  Kareem Ayoub and
                  Megha Goel and
                  Clemens Meyer and
                  Gregory Thornton and
                  Zhen Yang and
                  Henryk Michalewski and
                  Zaheer Abbas and
                  Nathan Schucher and
                  Ankesh Anand and
                  Richard Ives and
                  James Keeling and
                  Karel Lenc and
                  Salem Haykal and
                  Siamak Shakeri and
                  Pranav Shyam and
                  Aakanksha Chowdhery and
                  Roman Ring and
                  Stephen Spencer and
                  Eren Sezener and
                  et al.},
  title        = {Gemini 1.5: Unlocking multimodal understanding across millions of
                  tokens of context},
  journal      = {CoRR},
  volume       = {abs/2403.05530},
  year         = {2024}
}

@article{internvl2,
  author       = {Zhe Chen and
                  Weiyun Wang and
                  Hao Tian and
                  Shenglong Ye and
                  Zhangwei Gao and
                  Erfei Cui and
                  Wenwen Tong and
                  Kongzhi Hu and
                  Jiapeng Luo and
                  Zheng Ma and
                  Ji Ma and
                  Jiaqi Wang and
                  Xiaoyi Dong and
                  Hang Yan and
                  Hewei Guo and
                  Conghui He and
                  Botian Shi and
                  Zhenjiang Jin and
                  Chao Xu and
                  Bin Wang and
                  Xingjian Wei and
                  Wei Li and
                  Wenjian Zhang and
                  Bo Zhang and
                  Pinlong Cai and
                  Licheng Wen and
                  Xiangchao Yan and
                  Min Dou and
                  Lewei Lu and
                  Xizhou Zhu and
                  Tong Lu and
                  Dahua Lin and
                  Yu Qiao and
                  Jifeng Dai and
                  Wenhai Wang},
  title        = {How Far Are We to GPT-4V? Closing the Gap to Commercial Multimodal
                  Models with Open-Source Suites},
  journal      = {CoRR},
  volume       = {abs/2404.16821},
  year         = {2024}
}

@article{videollama2,
  author       = {Zesen Cheng and
                  Sicong Leng and
                  Hang Zhang and
                  Yifei Xin and
                  Xin Li and
                  Guanzheng Chen and
                  Yongxin Zhu and
                  Wenqi Zhang and
                  Ziyang Luo and
                  Deli Zhao and
                  Lidong Bing},
  title        = {VideoLLaMA 2: Advancing Spatial-Temporal Modeling and Audio Understanding
                  in Video-LLMs},
  journal      = {CoRR},
  volume       = {abs/2406.07476},
  year         = {2024}
}

@article{longvu,
  title={LongVU: Spatiotemporal Adaptive Compression for Long Video-Language Understanding},
  author={Shen, Xiaoqian and Xiong, Yunyang and Zhao, Changsheng and Wu, Lemeng and Chen, Jun and Zhu, Chenchen and Liu, Zechun and Xiao, Fanyi and Varadarajan, Balakrishnan and Bordes, Florian and others},
  journal={arXiv preprint arXiv:2410.17434},
  year={2024}
}

@inproceedings{internvideo2,
  author       = {Wang, Yi and Li, Kunchang and Li, Xinhao and Yu, Jiashuo and He, Yinan and Chen, Guo and Pei, Baoqi and Zheng, Rongkun and Xu, Jilan and Wang, Zun and others},
  title        = {Internvideo2: Scaling video foundation models for multimodal video understanding},
  booktitle    = {{ECCV}},
  year         = {2024}
}

@article{internvideo,
  title={Internvideo: General video foundation models via generative and discriminative learning},
  author={Wang, Yi and Li, Kunchang and Li, Yizhuo and He, Yinan and Huang, Bingkun and Zhao, Zhiyu and Zhang, Hongjie and Xu, Jilan and Liu, Yi and Wang, Zun and others},
  journal={arXiv preprint arXiv:2212.03191},
  year={2022}
}

@inproceedings{ego4d,
  title={Ego4d: Around the world in 3,000 hours of egocentric video},
  author={Grauman, Kristen and Westbury, Andrew and Byrne, Eugene and Chavis, Zachary and Furnari, Antonino and Girdhar, Rohit and Hamburger, Jackson and Jiang, Hao and Liu, Miao and Liu, Xingyu and others},
  booktitle={Proceedings of the IEEE/CVF Conference on Computer Vision and Pattern Recognition},
  pages={18995--19012},
  year={2022}
}

@inproceedings{howto100m,
  title={Howto100m: Learning a text-video embedding by watching hundred million narrated video clips},
  author={Miech, Antoine and Zhukov, Dimitri and Alayrac, Jean-Baptiste and Tapaswi, Makarand and Laptev, Ivan and Sivic, Josef},
  booktitle={Proceedings of the IEEE/CVF international conference on computer vision},
  pages={2630--2640},
  year={2019}
}

@inproceedings{mscoco,
  title={Microsoft coco: Common objects in context},
  author={Lin, Tsung-Yi and Maire, Michael and Belongie, Serge and Hays, James and Perona, Pietro and Ramanan, Deva and Doll{\'a}r, Piotr and Zitnick, C Lawrence},
  booktitle={{ECCV}},
  pages={740--755},
  year={2014},
  organization={Springer}
}

@inproceedings{perceptiontest,
  title={Perception test: A diagnostic benchmark for multimodal video models},
  author={Patraucean, Viorica and Smaira, Lucas and Gupta, Ankush and Recasens, Adria and Markeeva, Larisa and Banarse, Dylan and Koppula, Skanda and Malinowski, Mateusz and Yang, Yi and Doersch, Carl and others},
  journal={Advances in Neural Information Processing Systems},
  volume={36},
  year={2024}
}

@inproceedings{yu2025vrbench,
  title={Vrbench: A benchmark for multi-step reasoning in long narrative videos},
  author={Yu, Jiashuo and Wu, Yue and Chu, Meng and Ren, Zhifei and Huang, Zizheng and Chu, Pei and Zhang, Ruijie and He, Yinan and Li, Qirui and Li, Songze and others},
  booktitle={Proceedings of the IEEE/CVF International Conference on Computer Vision},
  pages={21655--21666},
  year={2025}
}

@article{fu2026video,
  title={Video-MME-v2: Towards the Next Stage in Benchmarks for Comprehensive Video Understanding},
  author={Fu, Chaoyou and Yuan, Haozhi and Dong, Yuhao and Zhang, Yi-Fan and Shen, Yunhang and Hu, Xiaoxing and Li, Xueying and Su, Jinsen and Long, Chengwu and Xie, Xiaoyao and others},
  journal={arXiv preprint arXiv:2604.05015},
  year={2026}
}

@article{wang2024survey,
  title={A survey on large language model based autonomous agents},
  author={Wang, Lei and Ma, Chen and Feng, Xueyang and Zhang, Zeyu and Yang, Hao and Zhang, Jingsen and Chen, Zhiyuan and Tang, Jiakai and Chen, Xu and Lin, Yankai and others},
  journal={Frontiers of Computer Science},
  volume={18},
  number={6},
  pages={186345},
  year={2024},
  publisher={Springer}
}

@article{zhang2025survey,
  title={A survey on the memory mechanism of large language model-based agents},
  author={Zhang, Zeyu and Dai, Quanyu and Bo, Xiaohe and Ma, Chen and Li, Rui and Chen, Xu and Zhu, Jieming and Dong, Zhenhua and Wen, Ji-Rong},
  journal={ACM Transactions on Information Systems},
  volume={43},
  number={6},
  pages={1--47},
  year={2025},
  publisher={ACM New York, NY}
}

@article{masterman2024landscape,
  title={The landscape of emerging ai agent architectures for reasoning, planning, and tool calling: A survey},
  author={Masterman, Tula and Besen, Sandi and Sawtell, Mason and Chao, Alex},
  journal={arXiv preprint arXiv:2404.11584},
  year={2024}
}

@inproceedings{ning2025survey,
  title={A survey of webagents: Towards next-generation ai agents for web automation with large foundation models},
  author={Ning, Liangbo and Liang, Ziran and Jiang, Zhuohang and Qu, Haohao and Ding, Yujuan and Fan, Wenqi and Wei, Xiao-yong and Lin, Shanru and Liu, Hui and Yu, Philip S and others},
  booktitle={Proceedings of the 31st ACM SIGKDD Conference on Knowledge Discovery and Data Mining V. 2},
  pages={6140--6150},
  year={2025}
}

@article{ning2026code,
  title={Code as Agent Harness},
  author={Ning, Xuying and Tieu, Katherine and Fu, Dongqi and Wei, Tianxin and Li, Zihao and Bei, Yuanchen and Zou, Jiaru and Ai, Mengting and Liu, Zhining and Li, Ting-Wei and others},
  journal={arXiv preprint arXiv:2605.18747},
  year={2026}
}

@article{wang2023internvid,
  title={Internvid: A large-scale video-text dataset for multimodal understanding and generation},
  author={Wang, Yi and He, Yinan and Li, Yizhuo and Li, Kunchang and Yu, Jiashuo and Ma, Xin and Li, Xinhao and Chen, Guo and Chen, Xinyuan and Wang, Yaohui and others},
  journal={arXiv preprint arXiv:2307.06942},
  year={2023}
}

@article{li2024videochat,
  title={VideoChat-Flash: Hierarchical Compression for Long-Context Video Modeling},
  author={Li, Xinhao and Wang, Yi and Yu, Jiashuo and Zeng, Xiangyu and Zhu, Yuhan and Huang, Haian and Gao, Jianfei and Li, Kunchang and He, Yinan and Wang, Chenting and others},
  journal={arXiv preprint arXiv:2501.00574},
  year={2024}
}

@article{wang2025scireasoner,
  title={SciReasoner: Laying the Scientific Reasoning Ground Across Disciplines},
  author={Wang, Yizhou and Tang, Chen and Deng, Han and Xiao, Jiabei and Liu, Jiaqi and Wu, Jianyu and Yao, Jun and Li, Pengze and Su, Encheng and Wang, Lintao and others},
  journal={arXiv preprint arXiv:2509.21320},
  year={2025}
}

@inproceedings{kuckreja2024geochat,
  title={Geochat: Grounded large vision-language model for remote sensing},
  author={Kuckreja, Kartik and Danish, Muhammad Sohail and Naseer, Muzammal and Das, Abhijit and Khan, Salman and Khan, Fahad Shahbaz},
  booktitle={Proceedings of the IEEE/CVF conference on computer vision and pattern recognition},
  pages={27831--27840},
  year={2024}
}

@inproceedings{gui2025webcode2m,
  title={Webcode2m: A real-world dataset for code generation from webpage designs},
  author={Gui, Yi and Li, Zhen and Wan, Yao and Shi, Yemin and Zhang, Hongyu and Chen, Bohua and Su, Yi and Chen, Dongping and Wu, Siyuan and Zhou, Xing and others},
  booktitle={Proceedings of the ACM on Web Conference 2025},
  pages={1834--1845},
  year={2025}
}

@inproceedings{yu2024metamath,
  title={Metamath: Bootstrap your own mathematical questions for large language models},
  author={Yu, Longhui and Jiang, Weisen and Shi, Han and Yu, Jincheng and Liu, Zhengying and Zhang, Yu and Kwok, James and Li, Zhenguo and Weller, Adrian and Liu, Weiyang},
  booktitle={International Conference on Learning Representations},
  volume={2024},
  pages={45040--45061},
  year={2024}
}

@inproceedings{cheng2024seeclick,
  title={Seeclick: Harnessing gui grounding for advanced visual gui agents},
  author={Cheng, Kanzhi and Sun, Qiushi and Chu, Yougang and Xu, Fangzhi and YanTao, Li and Zhang, Jianbing and Wu, Zhiyong},
  booktitle={Proceedings of the 62nd Annual Meeting of the Association for Computational Linguistics (Volume 1: Long Papers)},
  pages={9313--9332},
  year={2024}
}

@article{hendrycks2020measuring,
  title={Measuring massive multitask language understanding},
  author={Hendrycks, Dan and Burns, Collin and Basart, Steven and Zou, Andy and Mazeika, Mantas and Song, Dawn and Steinhardt, Jacob},
  journal={arXiv preprint arXiv:2009.03300},
  year={2020}
}

@article{wei2024measuring,
  title={Measuring short-form factuality in large language models},
  author={Wei, Jason and Karina, Nguyen and Chung, Hyung Won and Jiao, Yunxin Joy and Papay, Spencer and Glaese, Amelia and Schulman, John and Fedus, William},
  journal={arXiv preprint arXiv:2411.04368},
  year={2024}
}

\clearpage
\appendix

\end{document}